\documentclass[journal]{IEEEtran}
\usepackage{courier}
\usepackage{amsmath}
\usepackage{amssymb}
\usepackage{amsfonts}
\usepackage{bbding}
\usepackage{balance}
\usepackage{indentfirst}
\usepackage{xcolor}
\usepackage{multirow}
\usepackage{float}
\usepackage{graphicx}
\usepackage{graphics}
\usepackage{subfigure}
\usepackage{float}
\usepackage{bigstrut}
\usepackage{makecell}
\usepackage{booktabs}
\usepackage{bibentry}
\usepackage{epstopdf}
\usepackage{wasysym}
\usepackage{tabularx}
\usepackage{comment}
\usepackage{algorithm}
\usepackage{algpseudocode}
\usepackage{pifont}
\usepackage{url}
\usepackage{placeins}

 \makeatletter
    \newcommand{\thickhline}{%
        \noalign {\ifnum 0=`}\fi \hrule height 1pt
        \futurelet \reserved@a \@xhline
    }
    \newcolumntype{"}{@{\vrule width 1pt}}
    \makeatother
\makeatletter
\let\ftype@table\ftype@figure
\makeatother

\let\oldFootnote\footnote
\newcommand\nextToken\relax

\renewcommand\footnote[1]{%
    \oldFootnote{#1}\futurelet\nextToken\isFootnote}

\newcommand\isFootnote{%
    \ifx\footnote\nextToken\textsuperscript{,}\fi}

\ifCLASSINFOpdf
\else
\fi

\begin{document}

\title{PVRED: A Position-Velocity Recurrent Encoder-Decoder for Human Motion Prediction}
\author{Hongsong~Wang, Jian Dong, Bin Cheng, 
        and~Jiashi~Feng 
\IEEEcompsocitemizethanks{
\IEEEcompsocthanksitem Corresponding author: Jian Dong. E-mail: timeflydj@gmail.com. 
\IEEEcompsocthanksitem This research is supported by the National Research Foundation, Singapore under its AI Singapore Programme (AISG Award No: AISG-100E-2019-035), Singapore National Research Foundation (“CogniVision” grant NRF-CRP20-2017-0003)
\IEEEcompsocthanksitem H.~Wang finished this work when he was a research fellow at National University of Singapore.
E-mail: hongsongsui@gmail.com.
\IEEEcompsocthanksitem J.~Dong is with the Intelligent Engineering Department, Qihoo 360. E-mail: timeflydj@gmail.com.
\IEEEcompsocthanksitem B.~Cheng is with the Machine Learning Group, Beijing Academy of Artificial Intelligence, Beijing, China. E-mail: chengbin@baai.ac.cn. 
\IEEEcompsocthanksitem J.~Feng is with the Department of Electrical and Computer Engineering, National University of Singapore.
E-mail: elefjia@nus.edu.sg. }
}

\markboth{JOURNAL OF LATEX CLASS FILES, ~Vol.~xx, No.~xx, xx~2017}%
{Shell \MakeLowercase{\textit{et al.}}: Bare Demo of IEEEtran.cls for IEEE Journals}

\maketitle

\begin{abstract}
Human motion prediction, which aims to predict future human poses given past poses, has recently seen increased interest. Many recent approaches are based on Recurrent Neural Networks (RNN) which model human poses with exponential maps. These approaches neglect the pose velocity as well as temporal relation of different poses, and tend to converge to the mean pose or fail to generate natural-looking poses. We therefore propose a novel Position-Velocity Recurrent Encoder-Decoder (PVRED) for human motion prediction, which makes full use of pose velocities and temporal positional information. A temporal position embedding method is presented and a Position-Velocity RNN (PVRNN) is proposed. We also emphasize the benefits of quaternion parameterization of poses and design a novel trainable Quaternion Transformation (QT) layer, which is combined with a robust loss function during training.
We provide quantitative results for both short-term prediction in the future 0.5 seconds and long-term prediction in the future 0.5 to 1 seconds. Experiments on several benchmarks show that our approach considerably outperforms the state-of-the-art methods. In addition, qualitative visualizations in the future 4 seconds show that our approach could predict future human-like and meaningful poses in very long time horizons.
Code is publicly available on GitHub: \textcolor{red}{https://github.com/hongsong-wang/PVRNN}.
\end{abstract}

\maketitle
\IEEEdisplaynontitleabstractindextext
\IEEEpeerreviewmaketitle

\section{Introduction}\label{sec:introduction}
\IEEEPARstart{U}{nderstanding} and predicting human motion dynamics is an important topic in computer vision.
Human motion prediction aims to predict the future human motion dynamics given the past motion data. It has various applications including human-robot interaction~\cite{ayusawa2017motion}, augmented reality~\cite{villegas2018neural}, animation~\cite{holden2016deep}, etc. The temporal  changes of human poses show motion dynamics of the whole body. One common task of this problem is to forecast the most likely future 3D poses of a person by learning models from sequences of 3D poses. The task is challenging due to the non-rigid movement of articulated human body and the multimodal motion data, e.g., the sequence of an activity that consists of several submotions. The human pose is not so deterministic in the distant future, thus making it even more difficult for long-term ( i.e. more than 400 milliseconds) prediction.
\begin{figure}[!tbp]
\centering
\includegraphics[width=0.92\linewidth]{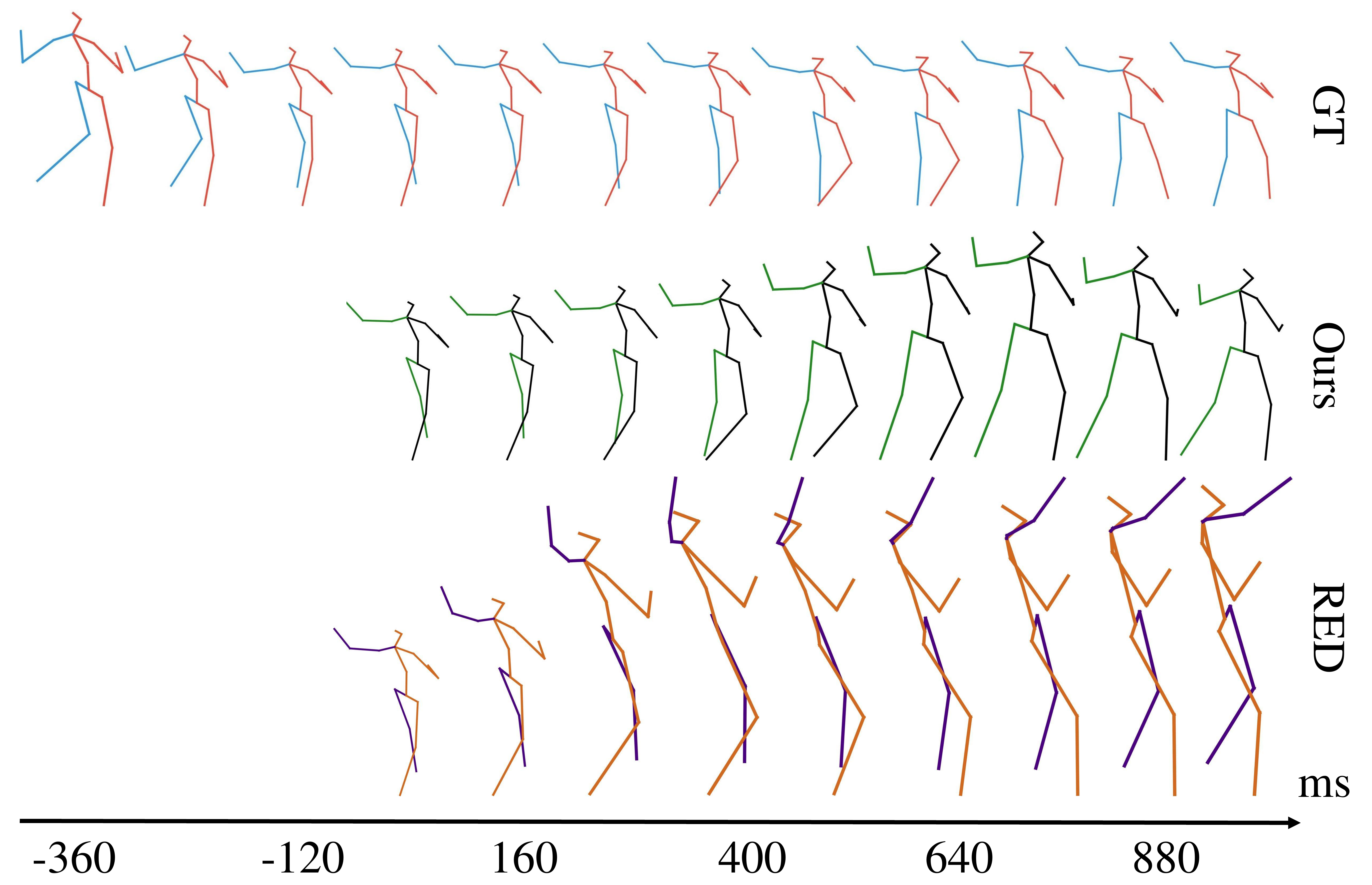}
\caption{Human motion prediction with different models. Given observations of past frames, the goal is to predict future frames of human poses in the next 1,000 milliseconds. We observe that the predictions from the proposed model are more accurate and natural than those of RED \cite{martinez2017human}. In this example, we show three observed poses and nine predicted poses. }
\label{fig:walking}
\end{figure}

Traditional approaches for learning dynamics of human motion mainly use probabilistic models including hidden Markov model~\cite{brand2000style}, linear dynamic system~\cite{pavlovic2001learning} and restricted Boltzmann machine~\cite{taylor2007modeling}. Prior knowledge about human motion is typically imposed and statistical models are used to constrain pose dynamics. Imposing physics based constraints is difficult and complex. Also, these approaches either generate unrealistic human motion or result in intractable estimation and inference problems.

Deep learning has also been successfully applied in human motion prediction. A family of methods based on Recurrent Neural Networks (RNN) are proposed, e.g., encoder-recurrent-decoder~\cite{fragkiadaki2015recurrent}, structural-RNN~\cite{jain2016structural} and RNN with de-noising autoencoder~\cite{ghosh2017learning}. These models learn structural and temporal dependencies from the training data and directly predict future poses. However, they tend to converge to the mean pose or fail to generate natural-looking poses. Martinez et al.~\cite{martinez2017human} proposed Recurrent Encoder-Decoder (RED) with residual connections that predicts velocities of body joint motion rather than poses. This approach mitigates the mean pose problem but still produces very inaccurate poses, especially for long-term predictions (see results of RED in Figure~\ref{fig:walking}).

Another disadvantage of RED based approaches is that the decoder RNN neglects temporal relations between poses of different frames due to autoregression based prediction. To address this issue, we aim to design a network which encodes temporal relative positions of different frames. This approach is partly inspired by the position embedding which is widely used in natural language processing~\cite{vaswani2017attention,gehring2017convolutional}.

For representing 3D human poses, the extensively used parameterization schemes include the Euler angle, exponential map and quaternion~\cite{grassia1998practical}. The exponential map is the most popularly adopted but it suffers from singularities (i.e., gimbal lock) and discontinuities of joint angles~\cite{grassia1998practical}. The quaternion parameterization is free of singularities and discontinuities of the representation, and would gain a more practical insight into human motion prediction. These advantages have been confirmed by the recently proposed QuaterNet~\cite{pavllo:quaternet:2018} which  employs quaternion to represent the input poses. However, QuaterNet abandons the raw input of exponential maps and also requires additional operations of preprocessing and postprocessing. An end-to-end network that makes full use of the quaternion parameterization is still absent.

Taking the pose velocities, relative position encoding and quaternion parameterization into consideration,  we propose a novel end-to-end trainable network, termed as Position-Velocity RED (PVRED), for human motion prediction. Different from previous methods~\cite{ghosh2017learning,tang2018long,chiu2018action,li2018convolutional}, the proposed network takes in three types of input: human poses, pose velocities and position embedding. To differentiate representations of adjacent and similar frames, we present an effective position embedding method based on sine and cosine functions of different frequencies to encode temporal positions of different frames. We design a Position-Velocity RNN (PVRNN) which constitutes the main part of PVRED. PVRNN takes in the three inputs and predicts pose velocities, which are then added to the previous poses to get the future poses. For the decoder, predictions of pose velocities and human poses are used as input for the next time stamp. To make use of the benefits of quaternion parameterization of 3D poses, we design a novel Quaternion Transformation (QT) layer to convert predicted poses from exponential maps to quaternion. The QT layer is embedded into the end-to-end trainable network. We also define a mean absolute error loss in the unit quaternion space to minimize the differences between the observed  and  predicted poses.

We make the following contributions. First, we propose a novel Position-Velocity Recurrent Encoder-Decoder (PVRED) for human motion prediction. Second, we first exploit temporal position embedding over frames while modeling the human pose sequences. Third, we design a novel Quaternion Transformation (QT) layer which takes advantages of quaternion parameterization of 3D poses for better pose prediction. Finally, our method obtains the state-of-the-art results for both short-term and long-term predictions of human motion with periodic actions as well as aperiodic actions.

\section{Related Work}\label{sec:related}
Predicting human motion dynamics is related to a range of research topics. Here we only review the previous works that are most related to ours.

\noindent \textbf{Human Motion Prediction}. Owing to the development of sequence-to-sequence models~\cite{cho2014learning,sutskever2014sequence}, several Recurrent Neural Networks (RNN) based approaches are proposed for human motion prediction~\cite{fragkiadaki2015recurrent,jain2016structural,martinez2017human,ghosh2017learning,gui2018adversarial,pavllo:quaternet:2018,gopalakrishnan2018a}.
Jain et al.~\cite{jain2016structural} developed a method for casting an arbitrary spatio-temporal graph as a fully differentiable and trainable RNN structure.
Martinez et al.~\cite{martinez2017human} presented a simple and effective baseline by adding a residual connection between the input and the output of each RNN cell.
Ghosh et al.~\cite{ghosh2017learning} combined a de-noising autoencoder with a 3-layer RNN to model the temporal aspects and recover the spatial structure of human pose.
Recently, Gopalakrishnan et al.~\cite{gopalakrishnan2018a} proposed a two-stage processing architecture which aids in generating planned trajectories.
Pavllo et al.~\cite{pavllo:quaternet:2018} designed an RNN architecture based on quaternions for rotation parameterization which shows the advantage of quaternions over exponential maps.

There are also some approaches beyond the RNN based ones~\cite{butepage2017deep,ruiz2018human,wang2019imitation,Aksan_2019_ICCV,li2018convolutional,wei2019motion,aksan2020spatio,li2020dynamic}. B{\"u}tepage et al.~\cite{butepage2017deep} proposed fully-connected networks with a bottleneck and directly fed the recent history poses to the model. Li et al.~\cite{li2018convolutional} utilized Convolutional Neural Networks (CNN) to learn to capture both invariant and dynamic information of human motion. Li et al.~\cite{li2020dynamic} propose a multiscale graph computational unit based on Graph Neural Networks (GNN) to model the internal relations of human poses. Mao et al.~\cite{wei2019motion} train a GNN to learn inter-joint dependencies based on discrete cosine transform coefficients of joints sequences. Recently, Aksan et al.~\cite{aksan2020spatio} apply Transformer to 3D human motion modeling and propose spatio-temporal transformer which models intra- and inter-joint dependencies.

These approaches mostly model the exponential map of human poses, and neglect temporal relations of poses at different time stamps. Our approach belongs to the paradigm with RNN and outperforms all the previous RNN-based approaches by taking full advantage of human poses, pose velocities and position embedding of different frames.

\vspace{1mm}
\noindent \textbf{Probabilistic Models}. Besides the deep learning based human motion prediction, there are some probabilistic models of human motion which can be applied in motion completion \cite{lehrmann2014efficient}, 3D action recognition~\cite{wang2017modeling,wang2018beyond}, etc.
Brand et al. \cite{brand2000style} used a hidden Markov model to generate new motion sequences of different styles.
Pavlovic et al. \cite{pavlovic2001learning} proposed switching linear dynamic system models to learn dynamic behaviour.
Sidenbladh et al. \cite{sidenbladh2002implicit} presented an implicit probabilistic model to provide a prior probability distribution over human motions.
Lehrmann et al. \cite{lehrmann2013non} introduced a non-parametric Bayesian network to generalize the prior of human pose with estimation of both graph structure and its local distribution.
Wang et al. \cite{wang2006gaussian} introduced a gaussian process dynamical model which comprises a latent space and a map from the latent space to an observation space.
Taylor et al. \cite{taylor2007modeling} used a conditional restricted Boltzmann machine to learn local constraints and global dynamics of human motion.
Lehrmann et al. \cite{lehrmann2014efficient} introduced the dynamic forest model which models human motion with an expressive Markov model.

These works exploit the low-dimensional representation of human motion with probabilistic models, while our approach obtains this representation using deep networks which could synthesize realistic motion sequences.

\vspace{1mm}
\noindent \textbf{Animation Synthesis}. Animation synthesis is a popular research topic in computer graphics. Here we only review deep learning based approaches. Taylor et al. \cite{taylor2009factored} present factored conditional restricted Boltzmann machines to generate motion of different styles. Holden et al. \cite{holden2015learning} learn a human motion manifold by a convolutional autoencoder and demonstrate applications including filling in missing motion data and interpolation of two motions. Holden et al. \cite{holden2016deep} propose fast and parallel feed forward neural networks for synthesizing character animation from control parameters. Different from human motion prediction, some high level instructions such as trajectories of the character should be additionally provided to synthesize character movements.

\section{Preliminaries}
In this section we revisit the RNN Encoder-Decoder (RED) paradigm and its application in human pose estimation as preliminaries.

\subsection{RNN Encoder-Decoder}
An RNN is a neural network that consists of a hidden state which operates on an input sequence of variable length. Given the input sequence $X$ with length $m$, i.e., $X = ({x_1}, \ldots ,{x_m})$, where ${x_t}$ is the input at time stamp $t$, the hidden state $h_t$ is updated by
\begin{equation}
{h_t} = f({h_{t - 1}},{x_t}) \label{eq:rnn}
\end{equation}
where $f$ is a non-linear activation function. The standard RNN suffers from the vanishing gradient problem, and some improved structures including Gated Recurrent Unit (GRU)~\cite{cho2014learning} and Long Short-Term Memory (LSTM)~\cite{hochreiter1997long} are designed by utilizing a  gating mechanism.

An RNN can learn a probability distribution over a sequence by being trained to predict the ground truth of the sequence. The output at each time stamp $t$ is the conditional distribution $p({x_t}|{x_1}, \ldots ,{x_{t - 1}})$. For example, Gaussian distribution can be used to model the distribution of ${x_t}$
\begin{equation}
p({x_t} = x|{x_1}, \ldots ,{x_{t - 1}}) = N(x|W{h_{t-1}} + b,{\sigma ^2}) \label{eq:guassin}
\end{equation}
where $W$ and $b$ are weight and bias parameters of linear regression, and ${\sigma ^2}$ is the variance of the distribution. The probability of the sequence $X$ is computed as
\begin{equation}
p(X) = \prod\limits_{t = 1}^m {p({x_t}|{x_1}, \ldots ,{x_{t - 1}})} \label{eq:seq_prob}
\end{equation}

The RNN Encoder-Decoder (RED)~\cite{cho2014learning,sutskever2014sequence} is a neural network architecture that consists of two RNNs, i.e., the encoder RNN and the decoder RNN. It first encodes an input sequence into a fixed-length vector representation and then decodes this vector into a new output sequence. Both the input sequence and the output sequence have a variable length. Suppose the input sequence is $X = ({x_1}, \ldots ,{x_n})$ and the output sequence is $Y = ({y_1}, \ldots ,{y_m})$.
From a probabilistic perspective, the distribution over the output sequence conditioned on the input sequence is $p({y_1}, \ldots ,{y_m}|{x_1}, \ldots ,{x_n})$.
For the encoder RNN, according to Equation (\ref{eq:rnn}), the encoded vector representation is the hidden state $h_n$ at the last time stamp of the encoder RNN. Similarly, for the decoder RNN, the hidden state ${\widetilde h_t}$ at time stamp $t$ can be updated based on ${\widetilde h_{t-1}}$ and $y_t$, where the hidden state ${\widetilde h_0}$ of the decoder RNN is $h_n$. With Equation (\ref{eq:guassin}) and (\ref{eq:seq_prob}), the probability of the output sequence $Y$ can be computed and the output vector ${y_t}$ can be predicted.

\begin{figure}[!tbp]
\centering
\includegraphics[width=0.85\linewidth]{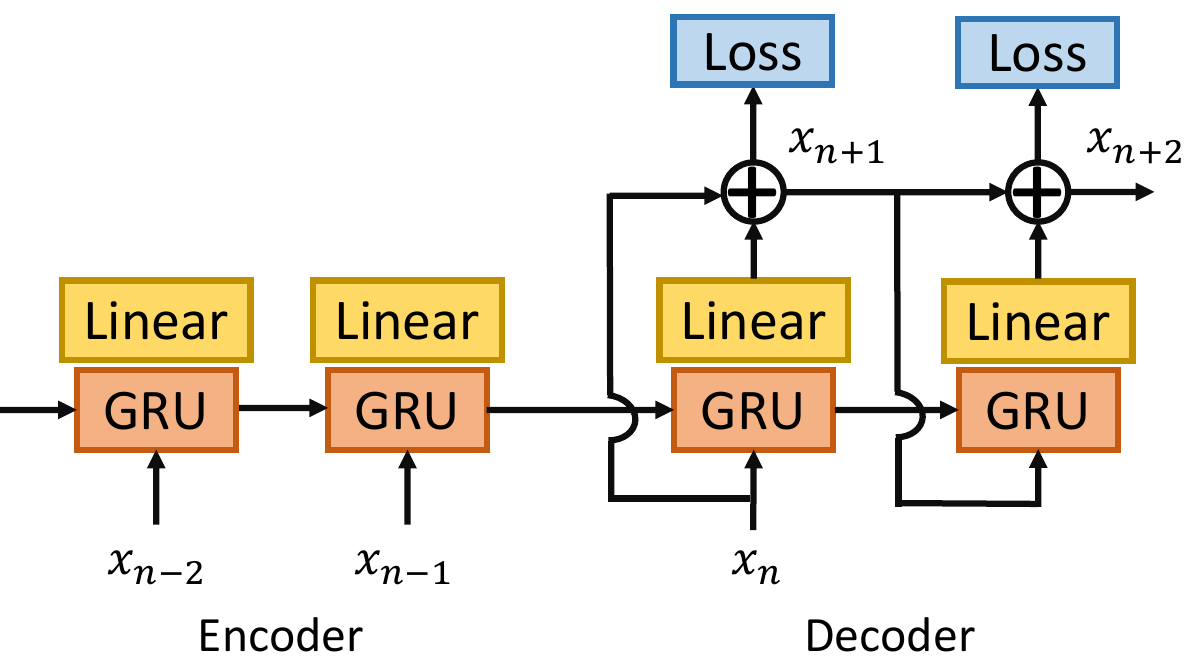}
\caption{A residual architecture of RNN Encoder-Decoder (RED)~\cite{martinez2017human}. For the encoder, the observed pose is the input at each frame. For the decoder, the input at a particular time stamp is its own previous prediction (except for the first time stamp). The decoder has a residual connection which forces the RNN to predict velocities.}
\label{fig:red}
\end{figure}

\subsection{RED for Human Motion Prediction}
Human motion prediction aims to forecast future poses based on a sequence of history 3D poses. With the human body considered as a skeleton of joints connected by bone segments, human poses can be described in two different ways, i.e., joint positions and joint angles. Forward kinematics refers to the calculation of joint positions from joint angles, and inverse kinematics denotes the computation of joint angles given a desired pose of joint locations. The exponential map parameterizes joint angles of 3D pose in three numerically stable Euclidean parameters. It offers several benefits such as robustness, simplicity, lack of explicit constraints and good modeling capabilities~\cite{grassia1998practical}.

For human motion prediction, joint angles of the exponential map are commonly used to describe human poses. The input sequence is $X = ({x_1}, \ldots ,{x_n})$, and the target sequence is $Y = ({x_{n + 1}}, \ldots ,{x_{n{\rm{ + }}m}})$, where $n$ and $m$ denote the lengths of the input and target sequences, respectively. For $1 \le i \le n + m$, the vector ${x_i} = {[e_{i1}^T, \ldots ,e_{ij}^T, \ldots ,e_{iJ}^T]^T}$, where $J$ is the number of joints of the human skeleton, and ${e_{ij}} = {[{e_{ijx}},{e_{ijy}},{e_{ijz}}]^\top}$ is the exponential map of the $j$-th joint of the $i$-th frame.

The RED can be applied to human motion prediction which can be regarded as a sequence-to-sequence learning problem. Suppose the given sequence $X$ has $n$ frames, i.e., $X = ({x_1}, \ldots ,{x_n})$, and the predicted future sequence $Y$ has $m$ frames, $Y = ({x_{n+1}}, \ldots ,{x_{n+m}})$. An RNN structure is used to model the input sequence $X$ and predict the output sequence $Y$. Based on the learned hidden state of the decoder RNN, the future poses can be predicted by using linear regression. Formally, for $j \in \{ 1, \ldots ,m\}$, the predicted pose at the future $j$-th frame is
\begin{equation}
x_{n+j} = W{h_{n + j - 1}} + b
\end{equation}
where $W$ and $b$ are weight and bias parameters, respectively, and when $j=1$, $h_n$ is the hidden state of the encoder RNN at the last time stamp.

While predicting future poses, the mean squared error loss is usually used to train the RED. The minimized loss function of a training sequence is
\begin{equation}
L = \frac{1}{m}\sum\limits_{j = 1}^m {{{\left\| {{y_{n + j}} - {x_{n + j}}} \right\|}_2}}
\end{equation}
where $y_{n + j}$ is the ground truth pose at time stamp $(n+j)$.

One good structure of RED for human motion prediction is illustrated in Figure~\ref{fig:red}, where the decoder RNN has a residual connection \cite{martinez2017human}. The predicted pose at the future $j$-th frame is
\begin{equation}
x_{n+j} = x_{n +j - 1} + W{h_{n + j - 1}} + b
\end{equation}
where $j \in \{1, \ldots ,m\}$, and when $j=1$, $x_n$ is the given pose of the input sequence at the last time stamp.

For human motion prediction, several good practices are exploited with the RED structure~\cite{martinez2017human}. For example, one layer of GRU~\cite{cho2014learning} is computationally inexpensive and achieves very competitive results. The LSTM~\cite{hochreiter1997long} is inferior to the GRU. Parameter sharing between the encoder RNN and the decoder RNN accelerates convergence. The residual connection ensures continuities between the conditioned sequence and prediction which could improve performance. We feed the prediction instead of the ground truth at each time stamp to the decoder RNN during training.

\section{Position-Velocity RED}
\begin{figure}[!tbp]
\centering
\includegraphics[width=0.85\linewidth]{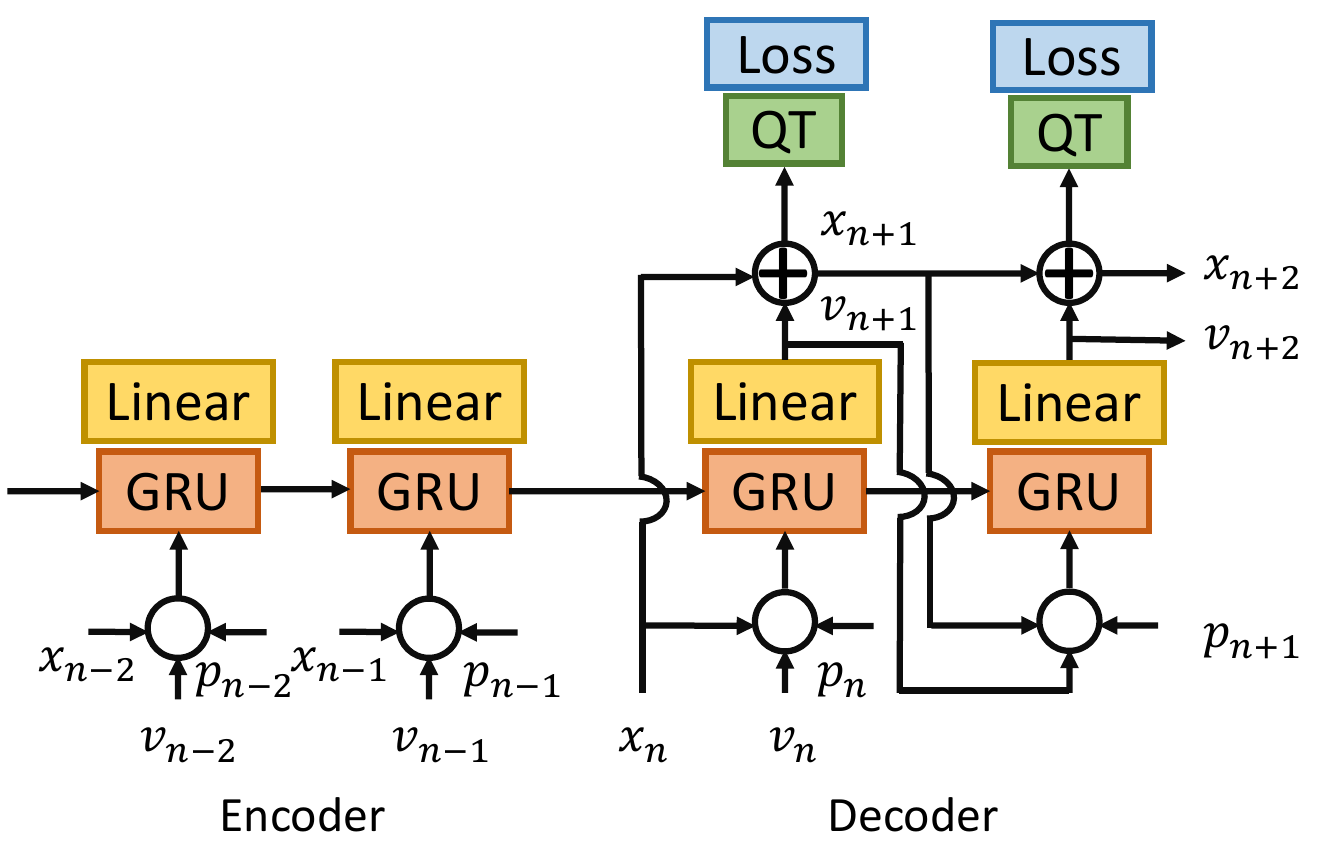}
\caption{The structure of the Position-Velocity Recurrent Encoder-Decoder (PVRED). Both the encoder and decoder have three types of input: poses $x_n$, velocities $v_n$ and positions $p_n$. QT denotes the proposed Quaternion Transformation layer.}
\label{fig:model}
\end{figure}

Based on the RED model, we propose a Position-Velocity Recurrent Encoder-Decoder (PVRED) which is shown in Figure~\ref{fig:model}. Different from RED, our PVRED takes in human poses, pose velocities and position embedding. The encoder takes in the three inputs at each time stamp, and derives the initial hidden state of the decoder from the given sequence. The decoder first predicts velocities of the next frame, and predicts corresponding poses with a residual connection. The predicted pose velocities and human poses are considered as the input of the decoder at the next time stamp. We also design a Quaternion Transformation (QT) layer and define a robust loss function of human poses in a unit quaternion space.

\subsection{Position Embedding} \label{sec:pos_emb}
While it is a common practice to incorporate position embedding in many natural language processing tasks~\cite{vaswani2017attention,gehring2017convolutional}, temporal position information is seldom used for computer vision tasks. The positional information encourages the model to learn more discriminative representations as it differentiates the representations of similar poses at different time stamps. It also has the potential to alleviate the mean pose problem that the predicted poses converge to an undesired mean pose.

Position embedding is to encode the absolute temporal positions of different frames into a real-valued vector which conveys the relative position information. One simple method is to use one-hot vector where the encoded vector is all zero values except for the index of the current frame, which is marked with one. The one-hot embedding is not flexible for encoding the sequence of a variable length. Inspired by the work~\cite{vaswani2017attention}, we use sine and cosine functions of different frequencies to encode the relative or absolute positions.

Assume that the given sequence has $n$ frames. We aim to predict the future $m$ frames. For a time stamp t, $t \in \{ 1, \ldots ,n, \ldots ,n + m\}$, the position embedding $p_t$ is expressed as
\begin{equation}
\begin{array}{l}
{p_t}(2i) = \sin(t/{10000^{2i/{d^p}}}),\\
{p_t}(2i - 1) = \cos (t/{10000^{2i/{d^p}}}),
\end{array}\label{eq:postion_emb}
\end{equation}
where $d^p$ is the embedding dimension, $i$ is the index, and $1 \le i \le \left\lceil {{{{d^p}} \mathord{\left/{\vphantom {{{d^p}} 2}} \right. \kern-\nulldelimiterspace} 2}} \right\rceil$. In~\cite{vaswani2017attention}, the dimension of position embedding is equal to the word embedding dimension to make the summation operation possible, and the final result is not sensitive to this hyper-parameter. Similarly for human motion prediction, $d^p$ is assigned the same dimension of coordinates of human pose to make some computational operations feasible. 

Each dimension of the positional embedding is a sinusoid. The wavelengths form a geometric progression from $2\pi$ to $10000 \cdot 2\pi$. For any fixed offset $k$, $p_{t+k}$ can be represented as a linear function of $p_t$. Therefore, the sinusoid embedding method allows the model to learn to attend by relative positions and predict natural-looking poses at different time intervals. It also allows this model to extrapolate to sequences of variable lengths during training.

\subsection{Position-Velocity RNN} \label{sec:pos_vel_rnn}
Given the input of human pose $x_t$ at each time stamp $t$, we consider the time derivative of $x_t$, i.e., the velocity of human poses. It is easy to preserve motion continuities in terms of velocity as it directly measures human motion. Following the good practice of previous works \cite{martinez2017human,gui2018adversarial,pavllo:quaternet:2018}, we adopt GRU as the unit cell, which has a reset gate $r_t$ and an update gate $z_t$. We combine velocity and position embedding with the input of human poses, and design a Position-Velocity RNN (PVRNN) to predict a sequence of human poses.

The proposed PVRNN has three inputs: human pose $x_t$, pose velocity $v_t$, and position embedding $p_t$. The hidden state $h_t$ at time stamp $t$ is computed as
\begin{equation}
\begin{array}{l}
{z_t} = \sigma ({U_x^z}{x_t} + {U_v^z}{v_t} + {U_p^z}{p_t} + {W^z}{h_{t - 1}})\\
{r_t} = \sigma ({U_x^r}{x_t} + {U_v^r}{v_t} + {U_p^r}{p_t} + {W^r}{h_{t - 1}})\\
{\tilde {h_t}}  = \tanh ({U_x^h}{x_t} + {U_v^h}{v_t} + {U_p^h}{p_t} + {W^h}({r_t} \circ {h_{t - 1}}))\\
{h_t} = (1 - {z_t}) \circ {h_{t - 1}} + {z_t} \circ {\tilde {h_t}}
\end{array}\label{eq:pvrnn}
\end{equation}
where $\circ$ denotes the Hadamard product, and variables $U$ (e.g., $U_x^z$, $U_v^z$) and $W$ (e.g., $W^z$) are weight matrices.

Given a sequence of history poses $X = ({x_1}, \ldots ,{x_n})$, we aim to predict future poses $Y = ({x_{n+1}}, \ldots ,{x_{n+m}})$ of the next $m$ time stamps. To estimate the future pose, the velocity is first predicted based on the hidden state and then added to the previous pose. Mathematically,
\begin{equation} \label{eq:our_pred}
\begin{array}{l}
{v_{n + j -1}} = W{h_{n + j -1}} + b\\
{x_{n + j}} = {x_{n + j -1}} + {v_{n + j -1}}
\end{array}
\end{equation}
where $W$ and $b$ are weights and bias parameters, respectively. For the decoder, $j \in \{ 1, \ldots ,m\}$, and when $j=0$, $h_n$ is the last time stamp of the PVRNN encoder.

\subsection{Quaternion Transformation} \label{QuatTransf}
For human motion prediction, human poses are mostly described by joint rotations using the exponential maps. The exponential map describes the axis and magnitude of a three DOF rotation, and is numerically stable. Despite the many advantages, the exponential map suffers from singularities (i.e., gimbal lock) and discontinuities in three-dimensional ${\mathbb{R}^3}$ of radius $2n\pi$ ($n = \{ 1,2 \ldots \}$)~\cite{grassia1998practical}. The 3D rotations can also be parameterized by unit-length quaternions in ${\mathbb{R}^4}$. The quaternions can get rid of singularities and discontinuities, and the multiplication operator in the quaternion space corresponds to matrix multiplication of rotation matrices. The recent work~\cite{pavllo:quaternet:2018} converts the raw input of exponential maps into quaternions and uses RED to predict the future joint rotations with quaternions as well. To enforce the unit length of quaternions, an explicit normalization layer is added to their network. This approach requires additional operations of preprocessing and postprocessing and is not end-to-end trainable.

Like most previous works~\cite{jain2016structural,martinez2017human,ghosh2017learning,li2018convolutional,gui2018adversarial}, we use the exponential map of joint rotations as the input of the proposed network. To enjoy the benefits of quaternion parameterization, we design a novel Quaternion Transformation (QT) layer to convert the predicted pose from exponential maps to quaternion. The QT layer could be embedded into the end-to-end trainable network. Assume that the human body has $J$ joints, and $e_j$ denotes the exponential map of joint $j$. The predicted or the ground truth pose at a particular time stamp is $x = {[e_1^\intercal, \ldots ,e_i^\intercal, \ldots ,e_J^\intercal]^\intercal}$. For simplicity, we use $e$ to denote $e_j$, which is a three-dimensional vector. The QT layer transforms $e$ into a four-dimensional vector $q$:
\begin{equation}
q(i) = \left\{ \begin{array}{ll}
\cos(0.5{\left\| e \right\|_2}) & i = 1\\
\frac{{\sin(0.5{{\left\| e \right\|}_2})}}{{{{\left\| e \right\|}_2}}} \cdot e(i - 1) & i \ge 2
\end{array} \right.
\end{equation}
where $q$ denotes the corresponding joint rotations in terms of quaternions, and ${{{\left\| e \right\|}_2}}$ is the $L^2$-norm of the vector $e$.

During backpropagation, the derivative of $q$ with respect to $e$ is the Jacobian matrix with dimensions $4 \times 3$, which is
\begin{equation}
\frac{{\partial q}}{{\partial e}} = \left[ \begin{array}{l}
\sin(0.5{\left\| e \right\|_2}) \cdot {\hat e^\intercal}\\
0.5\cos(0.5{\left\| e \right\|_2}) \cdot E + \frac{{\sin(0.5{{\left\| e \right\|}_2})}}{{{{\left\| e \right\|}_2}}}({I_3} - E)
\end{array} \right]
\end{equation}
where $I_3$ is the $3 \times 3$ identity matrix, and
\begin{equation}
\begin{array}{l}
\hat e = \frac{e}{{{{\left\| e \right\|}_2}}}\\
E = \hat e \otimes \hat e
\end{array}
\end{equation}
where $\hat e$ is the normalized vector of $e$, $\otimes$ denotes the outer product of two vectors, and $E$ is a $3 \times 3$ matrix.

\subsection{Training}
To train the proposed network, we aim to define a loss function in the unit quaternion space. The objective is to minimize the differences between observed poses and predicted poses while keeping the unit length of quaternion representations. The loss function should be robust against outliers of inaccurate predictions.
We define a novel training loss as
\begin{equation}
L = \frac{1}{m}\sum\limits_{j = 1}^m {{{\left\| {g({y_{n + j}}) - g({x_{n + j}})} \right\|}_1}}
\label{eq:loss}
\end{equation}
where $g$ denotes the Quaternion Transformation (QT), $x_{n + j}$ is the predicted pose of Equation~(\ref{eq:our_pred}), $y_{n + j}$ is the ground truth pose, and $\left\|\cdot\right\|_1$ is the mean absolute error. The detailed training process is summarized in Algorithm \ref{algoritm1}.

\renewcommand{\algorithmicrequire}{\textbf{Input:}}
\renewcommand{\algorithmicensure}{\textbf{Output:}}
\scalebox{0.95}{
\begin{minipage}{1\linewidth}
\begin{algorithm}[H]
\caption{Training process of the PVRED.}
\label{algoritm1}
\begin{algorithmic}[1]
\Require The observed sequence $X = ({x_1}, \ldots ,{x_n})$.
\Require The future sequence $Y = ({y_{n+1}}, \ldots ,{y_{n+m}})$.
\State Calculate the time derivative of $X$ to obtain the velocity sequence $({v_1}, \ldots ,{v_n})$ of the observed $n$ frames.
\State Obtain the temporal position embedding sequence $({p_1}, \ldots ,{p_n}, \ldots ,{p_{n+m}})$ of all the $(n+m)$ frames based on Equation (\ref{eq:postion_emb}).
\For{\(t = 1, \cdots ,n-1\) }
\State Calculate the hidden state $h_t$ of the encoder RNN based on Equation (\ref{eq:pvrnn}).
\EndFor
\State The predicted sequence Z $\leftarrow \{ \}$.
\For{\(t = n, \cdots ,n+m-1\) }
\State Calculate the hidden state $h_t$ of the decoder RNN based on Equation (\ref{eq:pvrnn}).
\State Predict the future frame $x_{t+1}$ based on Equation (\ref{eq:our_pred}).
\State Z.insert(\(x_{t+1}\))
\State Calculate the next future velocity $v_{t+1} \leftarrow x_{t+1} - x_t$.
\EndFor
\State Perform quaternion transformation for both the predicted sequence $({x_{n+1}}, \ldots ,{x_{n+m}})$ and the ground truth sequence $({y_{n+1}}, \ldots ,{y_{n+m}})$.
\State Compute the training loss $L$ based on Equation (\ref{eq:loss}).
\State Perform the backward pass by minimizing the loss $L$.
\Ensure The updated network.
\end{algorithmic}
\label{algoritm1}
\end{algorithm}
\end{minipage}
}
\vspace{1mm}

During testing, the QT layer and the loss function are discarded. The proposed network takes in human poses, pose velocities and position embedding and predicts future poses. Both human poses and pose velocities are represented by the original exponential map.

\section{Experiments}
We validate our approach on two important benchmarks: Human 3.6M dataset~\cite{ionescu2014human3} and CMU Motion Capture dataset\footnote{\url{http://mocap.cs.cmu.edu/}}. Comparisons of our method with the state-of-the-arts are performed and ablation analyses are provided.

\subsection{Datasets}
\noindent \textbf{Human3.6M}. The Human 3.6M dataset~\cite{ionescu2014human3} is a large-scale publicly available dataset with 3.6 million accurate 3D poses. Each 3D pose has 32 joints. It is recorded by a Vicon motion capture system, and consists of 15 activities. Both cyclic motions such as walking and non-cyclic motions such as smoking are included. The activities are conducted by seven different subjects, and each subject performs two trials for each activity. The dataset is challenging and widely used in human motion analysis due to large pose variations.
We follow the standard experimental setup~\cite{fragkiadaki2015recurrent,jain2016structural,martinez2017human}. The sequences are down sampled by two to obtain a frame rate of 25fps. The sequences of the subject indexed five are used for testing and the other sequences are used for training. The Euclidean distance between predictions and the ground truth in terms of Euler angle is measured, and the test errors are averaged across 8 different seed clips.

\vspace{1mm}
\noindent \textbf{CMU Motion Capture}. The CMU Motion Capture dataset is a large dataset which provides 3D pose data of 144 different subjects. It contains a large spectrum of movements including everyday movements such as walking and running as well as sport movements such as climbing and dancing. Each pose has 38 joints for this dataset. Similar to~\cite{li2018convolutional}, we choose actions for human motion prediction based on below criteria. We select single person actions, and remove two person interactions and the composition of several atomic actions. We also exclude the categories which do not provide enough training data. The sequences are down sampled to satisfy the frame rate of 25fps. We use the same train/test split as~\cite{li2018convolutional}, and calculate the Euclidean distance between predictions and the ground truth in terms of Euler angle. To make the results more stable, we report the averaged distance across 80 sampled seed clips.

\subsection{Implementation Details}
Similar to previous works~\cite{martinez2017human,pavllo:quaternet:2018,li2018convolutional}, we train a model by using data of all actions, and test the predicted error for each of the selected actions. The action label is not used and the proposed model is action-agnostic. Some works~\cite{martinez2017human,li2018convolutional} preprocess data by subtracting the mean pose and dividing the standard deviation. We focus on end-to-end training and do not normalize the raw data. Unless otherwise specified, the given past sequence has 50 frames (two seconds), and the predicted future sequence has 25 frames (one second). During training, we uniformly sample clips of a fixed length from the training data. The numbers of hidden units of RNN are 1,024 and 512 for the Human3.6M and CMU datasets, respectively. We set the dimension of position embedding the same as that of the original pose.

During training, dropout with a rate of 0.2 is utilized when predicting future poses. We adopt the Adam optimizer with a constant learning rate of 0.0001. Batch training is used with a mini-batch size of 128. The maximum number of training epochs is 20,000. Our implementation is based on PyTorch, and the code is available on GitHub: \textcolor{red}{https://github.com/hongsong-wang/PVRNN}.

\subsection{Evaluation on Human3.6M}
\begin{table*}[!htb]
  \centering
  \caption{Short-term prediction error on the Human3.6M dataset. The result is the mean angle error measured at \{80, 160, 320, 400\} milliseconds after the seed motion. The methods marked with * are based on Graph Neural Networks (GNN).}
    \begin{tabular}{l|cccc|cccc|cccc|cccc}
      \thickhline
    \multirow{2}[0]{*}{Milliseconds} & \multicolumn{4}{c|}{Walking}   & \multicolumn{4}{c|}{Eating}    & \multicolumn{4}{c|}{Smoking}   & \multicolumn{4}{c}{Discussion} \\
     \cline{2-17}
          & 80    & 160   & 320   & 400   & 80    & 160   & 320   & 400   & 80    & 160   & 320   & 400   & 80    & 160   & 320   & 400 \\
    \hline
    ERD \cite{fragkiadaki2015recurrent}   & 1.30  & 1.56  & 1.84  &  -- & 1.66  & 1.93  & 2.28  & -- & 2.34  & 2.74  & 3.73  &  --  & 2.67  & 2.97  & 3.23 & -- \\
    LSTM-3LR \cite{fragkiadaki2015recurrent} & 1.18  & 1.50  & 1.67  &  -- & 1.36  & 1.79  & 2.29  & -- & 2.05  & 2.34  & 3.10  &  -- & 2.25  & 2.33  & 2.45 &--  \\
    SRNN \cite{jain2016structural} & 1.08  & 1.34  & 1.60  & -- & 1.35  & 1.71  & 2.12  &  -- & 1.90  & 2.30  & 2.90  &  --& 1.67  & 2.03  & 2.20  & -- \\
    DAE-LSTM \cite{ghosh2017learning} & 1.00  & 1.11  & 1.39  & -- & 1.31  & 1.49  & 1.86  &  -- & 0.92  & 1.03  & 1.15  & -- & 1.11  & 1.20  & 1.38  & -- \\
    Zero-velocity \cite{martinez2017human} & 0.39  & 0.68  & 0.99  & 1.15  & 0.27  & 0.48  & 0.73  & 0.86  & 0.26  & 0.48  & 0.97  & 0.95  & 0.31  & 0.67  & 0.94  & 1.04 \\
    Res GRU unsup. \cite{martinez2017human} & 0.27  & 0.47  & 0.70  & 0.78  & 0.25  & 0.43  & 0.71  & 0.87  & 0.33  & 0.61  & 1.04  & 1.19  & 0.31  & 0.69  & 1.03  & 1.12 \\
    Res GRU sup. \cite{martinez2017human} & 0.28  & 0.49  & 0.72  & 0.81  & 0.23  & 0.39  & 0.62  & 0.76  & 0.33  & 0.61  & 1.05  & 1.15  & 0.31  & 0.68  & 1.01  & 1.09 \\
RNN-MHU \cite{tang2018long} & 0.32 & 0.53 & 0.69 & 0.77 & -- & -- & -- & -- &  -- & -- & -- & -- &  0.31 & 0.66 & 0.93 & 1.00 \\
    AGED w/ geo \cite{gui2018adversarial} & 0.28  & 0.42  & 0.66  & 0.73  & 0.22  & 0.35  & 0.61  & 0.74  & 0.30  & 0.55  & 0.98  & 0.99  & 0.30  & 0.63  & 0.97  & 1.06 \\
    TP-RNN \cite{chiu2018action} & 0.25  & 0.41  & 0.58  & 0.65  & 0.20  & 0.33  & 0.53 & 0.67  & 0.26  & 0.47  & 0.88  & 0.90  & 0.30  & 0.66  & 0.96  & 1.04 \\
    RNN-SPL~\cite{Aksan_2019_ICCV} & 0.26 & 0.40 & 0.67 & 0.78 & 0.21 & 0.34 & 0.55 & 0.69 & 0.26 & 0.48 & 0.96 & 0.94 & 0.30 & 0.66 & 0.95 & 1.05 \\
    Conv Seq2Seq \cite{li2018convolutional} & 0.33  & 0.54  & 0.68  & 0.73  & 0.22  & 0.36  & 0.58  & 0.71  & 0.26  & 0.49  & 0.96  & 0.92  & 0.32  & 0.67  & 0.94  & 1.01 \\
    QuaterNet \cite{pavllo:quaternet:2018} & 0.21  & 0.34 & 0.56  & 0.62  & 0.20  & 0.35  & 0.58  & 0.70  & 0.25  & 0.47  & 0.93  & 0.90 & 0.26  & 0.60  & 0.85  & 0.93 \\
    ST-Transformer~\cite{li2020dynamic} & 0.21 & 0.36 & 0.58 & 0.63 & 0.17 & 0.30 & 0.49 & 0.60 & 0.22 & 0.43 & 0.88 & 0.82 & \textbf{0.19} & 0.52 & 0.79 & 0.88 \\ 
    DCT-GCN\textsuperscript{*}~\cite{wei2019motion} & 0.18 & 0.31 & 0.49 &  \textbf{0.56} & \textbf{0.16} & \textbf{0.29} & 0.50 & 0.62 & 0.22 & 0.41 & 0.86 & 0.80  & 0.20 & \textbf{0.51} & \textbf{0.77} & \textbf{0.85} \\
    DMGNN\textsuperscript{*}~\cite{li2020dynamic} & \textbf{0.18} & \textbf{0.31} & \textbf{0.49} & 0.58 &  0.17 & 0.30 & \textbf{0.49} & \textbf{0.59} & \textbf{0.21} & \textbf{0.39} & \textbf{0.81} & \textbf{0.77} & 0.26 & 0.65 & 0.92 & 0.99 \\
    \hline
    Ours  & 0.20 & 0.35  & 0.54 & 0.59 & 0.18 & 0.32 & 0.54  & 0.66 & 0.22 & 0.44 & \textbf{0.81} & 0.91  & 0.24 & 0.60 & 0.83 & 0.93 \\
      \thickhline
    \end{tabular} 
  \label{tab:h36}
  \end{table*}
\begin{table*}[hbt!]
	\centering
	\caption{Short-term prediction errors for the other actions on the Human 3.6M dataset.}
	\begin{tabular}{l|cccc|cccc|cccc|cccc}
		\thickhline
		\multirow{2}[0]{*}{Milliseconds} & \multicolumn{4}{c|}{Directions} & \multicolumn{4}{c|}{Greeting}  & \multicolumn{4}{c|}{Phoning}   & \multicolumn{4}{c}{Posing} \\
		\cline{2-17}
		& 80    & 160   & 320   & 400   & 80    & 160   & 320   & 400   & 80    & 160   & 320   & 400   & 80    & 160   & 320   & 400 \\
		\hline
		Zero-velocity \cite{martinez2017human} & 0.42  & 0.58  & 0.80  & 0.89  & 0.54  & 0.87  & 1.25  & 1.40  & 0.54  & 0.85  & 1.57  & 1.70  & \textbf{0.25} & \textbf{0.55} & \textbf{1.14} & \textbf{1.37} \\
		Moving avg. 2 \cite{martinez2017human} & 0.47  & 0.62  & 0.81  & 0.89  & 0.60  & 0.92  & 1.28  & 1.43  & 0.56  & 0.88  & 1.58  & 1.71  & 0.28  & 0.56  & 1.14  & 1.38 \\
		Res GRU \cite{martinez2017human} & 0.45  & 0.68  & 0.93  & 1.05  & 0.53  & 0.88  & 1.33  & 1.50  & 0.50  & 0.77  & \textbf{1.20} & \textbf{1.31} & 0.43  & 0.89  & 1.68  & 2.02 \\
		\hline
		Ours  & \textbf{0.31} & \textbf{0.42} & \textbf{0.66} & \textbf{0.72} & \textbf{0.40} & \textbf{0.66} & \textbf{1.00} & \textbf{1.13} & \textbf{0.45} & \textbf{0.69} & 1.26  & 1.34  & 0.26  & 0.62  & 1.19  & 1.42 \\
		\hline
		\multirow{2}[0]{*}{Milliseconds} & \multicolumn{4}{c|}{Purchases} & \multicolumn{4}{c|}{Sitting}   & \multicolumn{4}{c|}{Sitting down} & \multicolumn{4}{c}{Taking photo} \\
		\cline{2-17}
		& 80    & 160   & 320   & 400   & 80    & 160   & 320   & 400   & 80    & 160   & 320   & 400   & 80    & 160   & 320   & 400 \\
		\hline
		Zero-velocity \cite{martinez2017human} & 0.57  & 0.84  & 1.09  & 1.18  & 0.40  & 0.64  & 1.06  & 1.74  & 0.69  & 1.11  & 1.38  & 1.51  & 0.25  & 0.53  & 0.83  & 0.97 \\
		Moving avg. 2 \cite{martinez2017human} & 0.61  & 0.88  & 1.11  & 1.20  & 0.43  & 0.69  & 1.09  & 1.77  & 0.71  & 1.13  & 1.40  & 1.53  & 0.27  & 0.56  & 0.86  & 0.99 \\
		Res GRU \cite{martinez2017human} & 0.58  & 0.86  & 1.24  & 1.35  & 0.44  & 0.76  & 1.27  & 1.95  & \textbf{0.52} & 0.99  & 1.50  & 1.74  & 0.29  & 0.62  & 1.01  & 1.16 \\
		\hline
		Ours  & \textbf{0.47} & \textbf{0.71} & \textbf{1.05} & \textbf{1.10} & \textbf{0.30} & \textbf{0.47} & \textbf{0.84} & \textbf{1.56} & 0.58  & \textbf{0.70} & \textbf{1.03} & \textbf{1.19} & \textbf{0.17} & \textbf{0.40} & \textbf{0.66} & \textbf{0.79} \\
		\hline
		\multirow{2}[0]{*}{Milliseconds} & \multicolumn{4}{c|}{Waiting}   & \multicolumn{4}{c|}{Walking dog} & \multicolumn{4}{c|}{Walking together} & \multicolumn{4}{c}{Average} \\
		\cline{2-17}
		& 80    & 160   & 320   & 400   & 80    & 160   & 320   & 400   & 80    & 160   & 320   & 400   & 80    & 160   & 320   & 400 \\
		\hline
		Zero-velocity \cite{martinez2017human} & 0.33  & 0.66  & 1.26  & 1.53  & 0.59  & 0.94  & 1.39  & 1.54  & 0.34  & 0.67  & 0.95  & 1.00  & 0.41  & 0.71  & 1.09  & 1.26 \\
		Moving avg. 2 \cite{martinez2017human} & 0.36  & 0.68  & 1.28  & 1.55  & 0.62  & 0.97  & 1.40  & 1.53  & 0.37  & 0.71  & 0.97  & 1.02  & 0.44  & 0.74  & 1.11  & 1.28 \\
		Res GRU \cite{martinez2017human} & 0.34  & 0.67  & 1.17  & 1.35  & 0.52  & 0.85  & 1.29  & 1.48  & 0.30  & 0.60  & 0.87  & 0.95  & 0.41  & 0.72  & 1.14  & 1.33 \\
		\hline
		Ours  & \textbf{0.23} & \textbf{0.49} & \textbf{0.93} & \textbf{1.15} & \textbf{0.45} & \textbf{0.74} & \textbf{1.13} & \textbf{1.29} & \textbf{0.17} & \textbf{0.38} & \textbf{0.59} & \textbf{0.64} & \textbf{0.31} & \textbf{0.53} & \textbf{0.87} & \textbf{1.03} \\
		\thickhline
	\end{tabular} 
	\label{tab:h36_short_other}
\end{table*}

\begin{table*}[!htb]
  \centering
  \caption{Long-term prediction error on the Human3.6M dataset. The error is measured at \{560, 1000\} milliseconds after the seed motion. The methods marked with * are based on Graph Neural Networks (GNN).}
    \begin{tabular}{l|cc|cc|cc|cc}
     \thickhline
    \multirow{2}[0]{*}{Milliseconds} & \multicolumn{2}{c|}{Walking} & \multicolumn{2}{c|}{Eating} & \multicolumn{2}{c|}{Smoking} & \multicolumn{2}{c}{Discussion} \\
    \cline{2-9}
    & 560   & 1000  & 560   & 1000  & 560   & 1000  & 560   & 1000 \\
    \hline
    Zero-velocity \cite{martinez2017human} & 1.35  & 1.32  & 1.04  & 1.38  & 1.02  & 1.69  & 1.41  & 1.96 \\
    ERD \cite{fragkiadaki2015recurrent} & 2.00  & 2.38  & 2.36  & 2.41  & 3.68  & 3.82  & 3.47  & 2.92 \\
    LSTM-3LR \cite{fragkiadaki2015recurrent} & 1.81  & 2.20  & 2.49  & 2.82  & 3.24  & 3.42  & 2.48  & 2.93 \\
    SRNN  \cite{jain2016structural} & 1.90  & 2.13  & 2.28  & 2.58  & 3.21  & 3.23  & 2.39  & 2.43 \\
    DAE-LSTM \cite{ghosh2017learning} & 1.55  & 1.39  & 1.76  & 2.01  & 1.38  & 1.77  & 1.53  & 1.73 \\
    Res GRU sup. \cite{martinez2017human} & 0.93  & 1.03  & 0.95  & 1.08  & 1.25  & 1.50  & 1.43  & 1.69 \\
    AGED w/ geo \cite{gui2018adversarial} & 0.89  & 1.02  & 0.92  & \textbf{1.01} & 1.15  & 1.43  & 1.33  & 1.56 \\
    TP-RNN \cite{chiu2018action} &  0.74 & 0.77 & 0.84 & 1.14 & 0.98 &  1.66 & 1.39 &  1.74 \\
    Conv Seq2Seq \cite{li2018convolutional} & -- & 0.92 & -- & 1.24 & -- & 1.62 & -- & 1.86 \\
    DMGNN\textsuperscript{*}~\cite{li2020dynamic} & 0.66 & 0.75 & \textbf{0.74} & 1.14 & \textbf{0.74} & \textbf{1.14} & 1.33 & \textbf{1.45} \\
 \hline
    Ours  & \textbf{0.65} & \textbf{0.66} & 0.76 & 1.14  & 0.97 & 1.42 & \textbf{1.29} & 1.77 \\
    \thickhline
    \end{tabular} 
  \label{tab:h36_long}
\end{table*}
\begin{table*}[hbt!]
  \centering
  \caption{Long-term prediction errors for the other actions on the Human 3.6M dataset.}
  \begin{tabular}{l|cc|cc|cc|cc|cc|cc}
  \thickhline
    \multirow{2}[0]{*}{Milliseconds} & \multicolumn{2}{c|}{Directions} & \multicolumn{2}{c|}{Greeting} & \multicolumn{2}{c|}{Phoning} & \multicolumn{2}{c|}{Posing} & \multicolumn{2}{c|}{Purchases} & \multicolumn{2}{c}{Sitting} \\
    \cline{2-13}
          & 560   & 1000  & 560   & 1000  & 560   & 1000  & 560   & 1000  & 560   & 1000  & 560   & 1000 \\
    \hline
    Zero-velocity \cite{martinez2017human} & 1.04  & 1.51  & 1.63  & 1.72  & 1.94  & 2.23  & 1.79  & 2.81  & 1.57  & 2.45  & 1.78  & 2.07 \\
    Moving avg. 2 \cite{martinez2017human} & 1.04  & 1.51  & 1.65  & 1.73  & 1.95  & 2.24  & 1.79  & 2.81  & 1.60  & 2.48  & 1.80  & 2.09 \\
    Res GRU \cite{martinez2017human} & 1.15  & 1.64  & 1.82  & 2.14  & 1.55  & 2.05  & 2.39  & 2.85  & 1.76  & 2.58  & 2.11  & 2.60 \\
    \hline
    Ours  & \textbf{0.89} & \textbf{1.45} & \textbf{1.36} & \textbf{1.62} & \textbf{1.54} & \textbf{1.75} & \textbf{1.60} & \textbf{2.44} & \textbf{1.48} & \textbf{2.35} & \textbf{1.66} & \textbf{1.91} \\
     \hline
    \multirow{2}[0]{*}{Milliseconds} & \multicolumn{2}{c|}{Sitting down} & \multicolumn{2}{c|}{Taking photo} & \multicolumn{2}{c|}{Waiting} & \multicolumn{2}{c|}{Walking dog} & \multicolumn{2}{c|}{Walking together} & \multicolumn{2}{c}{Average} \\
    \cline{2-13}
          & 560   & 1000  & 560   & 1000  & 560   & 1000  & 560   & 1000  & 560   & 1000  & 560   & 1000 \\
    \hline
    Zero-velocity \cite{martinez2017human} & 1.67  & \textbf{2.05} & 1.07  & 1.31  & 1.97  & 2.71  & 1.71  & 1.86  & 1.10  & 1.47  & 1.48  & 1.90 \\
    Moving avg. 2 \cite{martinez2017human} & 1.69  & 2.07  & 1.09  & 1.33  & 1.98  & 2.72  & 1.69  & 1.83  & 1.11  & 1.48  & 1.49  & 1.90 \\
    Res GRU \cite{martinez2017human} & 2.08  & 2.87  & 1.33  & 1.64  & 1.64  & \textbf{2.22} & 1.66  & 1.92  & 1.14  & 1.61  & 1.57  & 2.04 \\
    \hline
    Ours  & \textbf{1.40} & 2.06  & \textbf{0.88} & \textbf{1.10} & \textbf{1.55} & 2.28  & \textbf{1.49} & \textbf{1.75} & \textbf{0.75} & \textbf{1.26} & \textbf{1.22} & \textbf{1.66} \\
    \thickhline
    \end{tabular}
  \label{tab:h36_long_other}
\end{table*}

\begin{figure*}[!htb]
\centering
\includegraphics[width=1.0\linewidth]{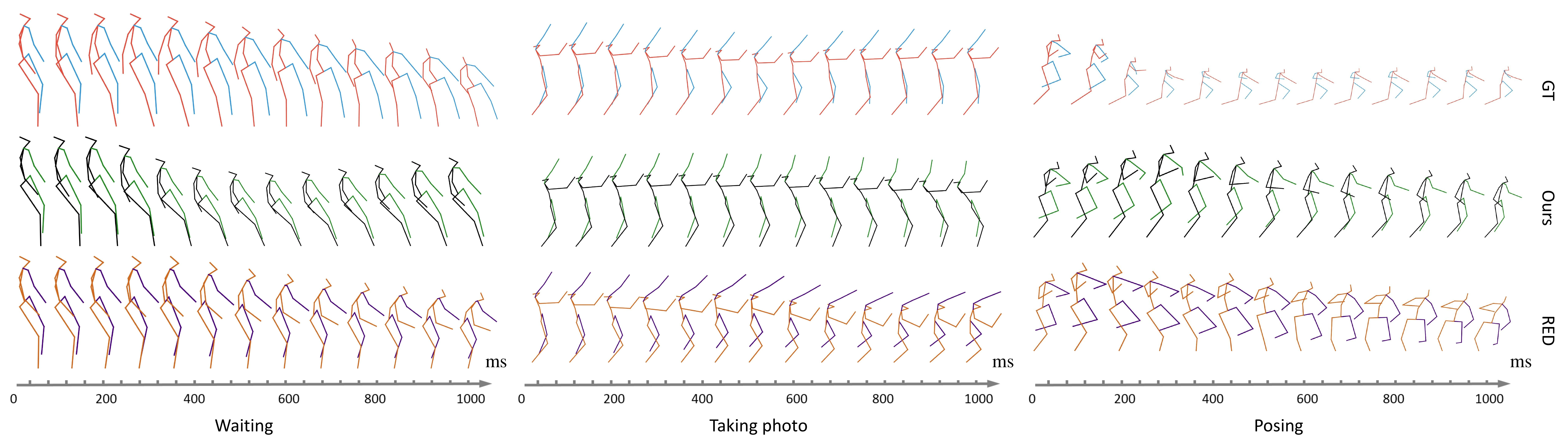}
\caption{Qualitative predicted poses in the future 1,000 milliseconds after the seed motion. Our predictions are less faraway from the ground-truth than predictions of RED, especially in the long term (e.g., more than 600 milliseconds). Best viewed in color with zoom.}
\label{fig:short_term_vis}
\end{figure*}
\begin{figure*}[!htb]
\centering
\includegraphics[width=1.0\linewidth]{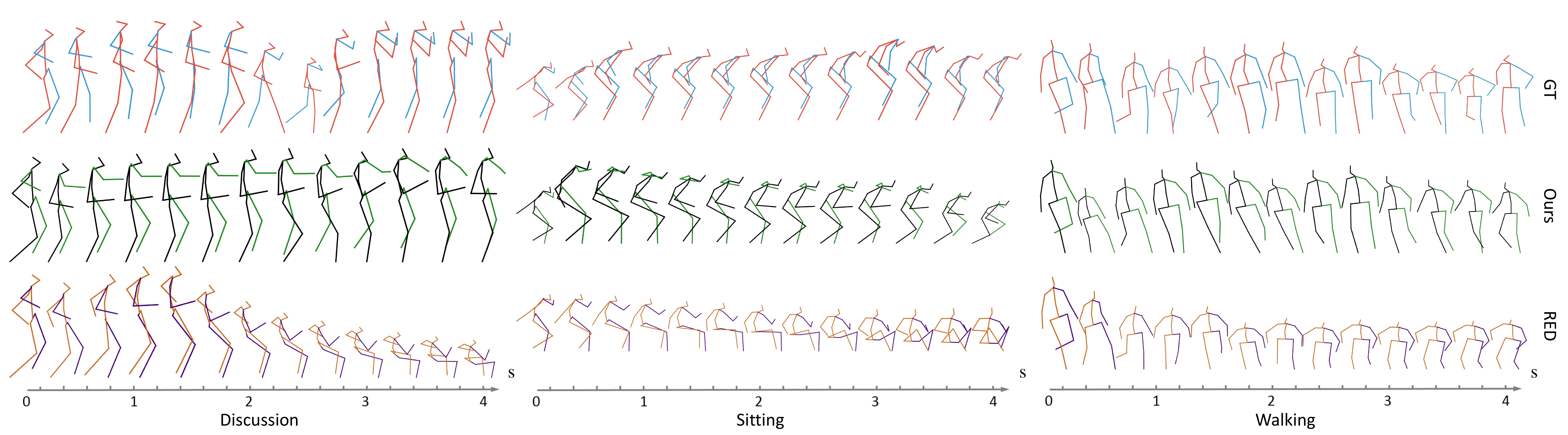}
\caption{Qualitative predicted poses in the future 4,000 milliseconds after the seed motion. RED predicts strange poses in long time horizons, while our predictions look plausible and show little difference with the real poses.}
\label{fig:long_term_vis}
\end{figure*}
\noindent \textbf{Short-Term Prediction}. Following previous convention~\cite{jain2016structural,gui2018adversarial,li2018convolutional}, we consider the prediction less than 500 milliseconds as short-term prediction. Within this time range, motion is almost deterministic and fairly predictable. In accordance with most previous works \cite{fragkiadaki2015recurrent,jain2016structural,martinez2017human,pavllo:quaternet:2018}, we consider four representative actions: walking, smoking, eating, and discussion. Walking is periodic and the other three are aperiodic. Table~\ref{tab:h36} compares prediction errors with previous approaches on the Human3.6M dataset. Among the RNN-based or CNN-based approaches, our approach yields the state-of-the-art performance for all actions at different time stamps. For example, for walking and at 80 milliseconds, our approach beats the strong baseline Residual RNN~\cite{martinez2017human} by more than 0.07, which is a significant margin when compared with the improvement of the contemporaneous approaches. Our method is also considerably superior to the recent quaternion based RNN~\cite{pavllo:quaternet:2018} which models and predicts human motion in the quaternions space, and the sequence-to-sequence model based on CNN~\cite{li2018convolutional}.
Since Graph Neural Networks (GNN) or Transformer have inherent advantages of modeling intra- and inter-joint dependencies over RNN, our results are slightly inferior to those of the few recent approaches based on GNN or Transformer.

Short-term prediction results of the other 11 actions and the averaged prediction errors across all actions are shown in Table~\ref{tab:h36_short_other}. Our approach beats the other state-of-the-art methods based on RNN in most cases. For the averaged performance across all actions, our prediction errors are 0.1, 0.19, 0.27 and 0.3 lower than those of \cite{martinez2017human} for the future 80, 160, 320 and 400 milliseconds, respectively. The amount of error reduction is evident for human motion prediction.

\vspace{1mm}
\noindent \textbf{Long-Term Prediction}. Motion prediction no less than 500 milliseconds is regarded as long-term prediction, which is more challenging than short-term prediction due to the stochastic nature and uncertainty of human motion. The results on the Human3.6M dataset are given in Table~\ref{tab:h36_long}. Our approach attains the best results nearly in most scenarios for both periodic actions and aperiodic actions. Specifically, for walking at 1,000 milliseconds, our approach decreases the reported lowest error, i.e., 0.92 of Conv Seq2Seq~\cite{li2018convolutional}, by 0.26, and beats the Residual RNN~\cite{martinez2017human} by 0.37. For predictions at 560 milliseconds, our predicted errors are 0.16, 0.18 and 0.04 lower than the best published results for eating, smoking, and discussion, respectively.
to make some computational operations feasible.
Table~\ref{tab:h36_long_other} shows long-term prediction results of the other actions on the Human3.6M dataset. Our approach outperforms all the other approaches by a considerable margin. Taking walking together as an example, our prediction error is 0.75 at 560 milliseconds, which is only 66\% of that of ~\cite{martinez2017human}.

\vspace{1mm}
\noindent \textbf{Visualization of Prediction}. We provide qualitative results by visualizing predicted poses of the test data, which are shown in Figure~\ref{fig:short_term_vis}. We observe that our approach mitigates the mean pose problem, and makes accurate short-term predictions and natural-looking long-term predictions. We also observe that predictions of RED (i.e., the Residual RNN~\cite{martinez2017human}) freeze to some mean poses and go faraway from the ground-truth in the long term. For example, predictions of RED for taking photo converge to a fixed pose at about 600 milliseconds, and the converged pose and the real pose vary considerably. In contrast, our predictions stay close to the ground-truth even in the future 1,000 milliseconds. More results of visualizations are in the videos \footnote{\url{https://www.youtube.com/watch?v=Gmvk8HsyzOc&t=5s}}\footnote{\url{https://www.youtube.com/watch?v=mECybzziYHM}}.

We also visualize predicted poses in very long time horizons. As the decoder RNN could generate sequences of variable lengths, we predict future human poses of the next 100 frames (four seconds) given past poses of 50 frames (two seconds). The same experimental settings and parameters are used except the length of predicted sequence. After training, we visualize predicted poses of the test samples. Figure~\ref{fig:long_term_vis} shows the results of representative actions, e.g., periodic walking and aperiodic sitting. We find that our approach could predict human-like and meaningful poses in the future 4,000 milliseconds, while predictions of RED quickly drift away to non-human-like poses.
As the recent Transformer-based model~\cite{aksan2020spatio} makes plausible and qualitative long-term predictions up to 15 seconds, we believe that the Transformer has great potential in human motion synthesis.

We also provide the predicted pose errors at 4000 milliseconds in Table~\ref{tab:h36_4000}. We compare our approach with the baselines such as LSTM~\cite{hochreiter1997long} and zero-velocity~\cite{martinez2017human}. We find that our approach dramatically beats the state-of-the-art method Res GRU~\cite{martinez2017human} and the LSTM baseline, while the simple zero-velocity baseline could achieve a close match with ours. The zero-velocity baseline constantly considers the future poses as the last pose of the observed sequence, and the predictions are meaningless for practical applications for human motion synthesis and animation synthesis. Since human motion in the future 4000 milliseconds is very uncertain, predicting human poses in such a long duration cannot be deterministic. Therefore, we only visualize predicted poses to show that our approach has the potential to synthesize meaningful and human-like poses in very long time horizons.

\begin{table}[htbp]
	\centering
	\caption{Human motion prediction in the future 4000 milliseconds for four representative actions on the Human 3.6M dataset.}
	\begin{tabular}{l|c|c|c|r}
	 \thickhline
		Method & \multicolumn{1}{c|}{Walking} & \multicolumn{1}{c|}{Eating} & \multicolumn{1}{c|}{Smoking} & \multicolumn{1}{c}{Discussion } \\
		\hline
		Zero-velocity~\cite{martinez2017human} & 1.57  & \textbf{2.08} & \textbf{2.27} & 2.44 \\
		LSTM  & 2.27  & 2.67  & 3.24  & 2.57 \\
		Res GRU~\cite{martinez2017human} & 2.65  & 2.61  & 2.95  & 2.75 \\
		\hline
		Ours  & \textbf{1.50} & 2.36  & 2.59  & \textbf{2.16} \\
	 \thickhline
	\end{tabular}
	\label{tab:h36_4000}
\end{table}

\subsection{Evaluation on CMU Motion Capture}
\begin{table*}[!htb]
  \centering
  \caption{Short-term prediction error on the CMU Motion Capture dataset. The results are averaged over 80 seed motion sequences for each activity on the test set. The methods marked with * are based on Graph Neural Networks (GNN).}
    \begin{tabular}{l|cccc|cccc|cccc|cccc}
    \thickhline
    \multirow{2}[0]{*}{Milliseconds} & \multicolumn{4}{c|}{Walking}   & \multicolumn{4}{c|}{Washing} & \multicolumn{4}{c|}{Basketball} & \multicolumn{4}{c}{Jumping} \\
    \cline{2-17}
          & 80    & 160   & 320   & 400   & 80    & 160   & 320   & 400   & 80    & 160   & 320   & 400   & 80    & 160   & 320   & 400 \\
    \hline
    Res GRU \cite{martinez2017human} & 0.29  & 0.45  & 0.66  & 0.73  & 0.34  & 0.66  & 1.02  & 1.13  & 0.42  & 0.73  & 1.20  & 1.35  & 0.63  & 0.91  & 1.44  & 1.67 \\
 Zero-velocity \cite{martinez2017human} & 0.30  & 0.50  & 0.80  & 0.93  & 0.33  & 0.53  & 0.89  & 1.03  & 0.48  & 0.85  & 1.47  & 1.71  & 0.46  & 0.68  & 1.22  & 1.44 \\
    Moving avg. 2 \cite{martinez2017human} & 0.32  & 0.51  & 0.82  & 0.94  & 0.35  & 0.56  & 0.91  & 1.05  & 0.52  & 0.90  & 1.51  & 1.74  & 0.49  & 0.72  & 1.25  & 1.46 \\
    QuaterNet \cite{pavllo:quaternet:2018} & 0.32 & 0.41 & 0.52 & 0.56 & 0.34 & 0.50 & 0.75 & 0.92 & 0.62 & 0.92 & 1.39 & 1.56 & 0.80 & 1.12 & 1.56 & 1.74 \\
    Conv Seq2Seq \cite{li2018convolutional} & 0.35  & 0.44  & 0.45  & 0.50  & 0.30  & 0.47  & 0.80  & 1.01  & 0.37  & 0.62  & 1.07  & 1.18  & \textbf{0.39} & \textbf{0.60} & 1.36  & 1.56 \\
    DMGNN\textsuperscript{*}~\cite{li2020dynamic} & 0.30 & \textbf{0.34} & 0.38 & 0.43 & \textbf{0.20} & \textbf{0.27} & \textbf{0.62} & 0.81 & \textbf{0.30} & \textbf{0.46} & \textbf{0.89} & \textbf{1.11} & 0.37 & 0.65 & 1.49 & 1.71 \\
 \hline
   Ours  & \textbf{0.28} & \textbf{0.34} & 0.41 & \textbf{0.43} & 0.25 & 0.37 & 0.67 & \textbf{0.81} & 0.36 & 0.56 & 0.95 & 1.13 & 0.46  & 0.65  & \textbf{1.14} & \textbf{1.34} \\
    \thickhline
    \end{tabular} 
  \label{tab:cmu_short}
\end{table*}
\begin{table*}[!htb]
  \centering
  \caption{Long-term prediction error on the CMU Motion Capture dataset. Our model consistently achieves the best performance.}
    \begin{tabular}{l|cc|cc|cc|cc}
    \thickhline
    \multirow{2}[0]{*}{Milliseconds} & \multicolumn{2}{c|}{Walking} & \multicolumn{2}{c|}{Washing} & \multicolumn{2}{c|}{Basketball} & \multicolumn{2}{c}{Jumping} \\ 
    \cline{2-9}
          & 560   & 1000  & 560   & 1000  & 560   & 1000  & 560   & 1000 \\
     \hline
    Res GRU \cite{martinez2017human} & 0.80  & 0.83  & 1.22  & 1.23  & 1.51  & 1.64  & 1.87  & 2.17 \\
   Zero-velocity \cite{martinez2017human} & 1.10  & 1.26  & 1.27  & 1.53  & 2.08  & 2.54  & 1.75  & 1.77 \\
    Moving avg. 2 \cite{martinez2017human} & 1.11  & 1.26  & 1.28  & 1.53  & 2.09  & 2.53  & 1.76  & 1.76 \\
    QuaterNet \cite{pavllo:quaternet:2018} & 0.61 & 0.72 & 1.10 & 1.23 & 1.71 & 1.81 & 1.89 & 2.09 \\
    Conv Seq2Seq \cite{li2018convolutional} & -- & 0.78  &  --   & 1.39  &  --   & 1.95  &   --   & 2.01 \\
    DMGNN\textsuperscript{*}~\cite{li2020dynamic} & -- & 0.60 & -- & \textbf{1.09} & -- & 1.66 & -- & 1.79 \\
 \hline
    Ours  & \textbf{0.47} & \textbf{0.53} & \textbf{1.02} & 1.20 & \textbf{1.41} & \textbf{1.61} & \textbf{1.57} & \textbf{1.75} \\
    \thickhline
    \end{tabular} 
  \label{tab:cmu_long}
\end{table*}
The CMU Motion Capture dataset is recently used for human motion prediction, and there are only a few reported results. We consider an agnostic zero-velocity baseline which constantly predicts the last observed frame~\cite{martinez2017human}, and the moving average baseline with a window size of two~\cite{martinez2017human}. The two baselines are simple but effective, and outperform many learning based approaches.

\noindent \textbf{Short-Term Prediction}. Similar to the Human3.6M dataset, we report results for four representative actions: walking, washing (washing window), basketball, and jumping. Jumping is aperiodic, and the other three are periodic. The results are summarized in Table~\ref{tab:cmu_short}. Our approach consistently outperforms the zero-velocity baseline as well as the strong baseline Residual RNN~\cite{martinez2017human}. In most cases, our approach exceeds the recent sequence-to-sequence model based on CNN~\cite{martinez2017human}.

\vspace{1mm}
\noindent \textbf{Long-Term Prediction}. The errors of long-term prediction are presented in Table~\ref{tab:cmu_long}. Our results are much better than those of the comparative approaches for both periodic actions and aperiodic actions. For walking, jumping and basketball, our predicted errors are 0.14, 0.32 and 0.30 lower than those of QuaterNet~\cite{pavllo:quaternet:2018} at 560 milliseconds, respectively. These experiments further confirm advantages of our approach for long-term predictions.

\subsection{Ablation Studies}
We run a number of ablations to analyze the proposed model. Without loss of generality, we only give the results on the Human3.6M dataset. The short-term prediction and long-term prediction are summarized in Table~\ref{tab:ablation} and Table~\ref{tab:ablation_long}, respectively. For simplicity, pose velocity, position embedding, and quaternion transformation are abbreviated as VEL, POS and QT, respectively. Different combinations of Vel, Pos and QT correspond to six variants of the proposed method, which are denoted by Var. 1-6, respectively. For example, Var. 1 refers to the approach that utilizes position embedding and quaternion transformation, and Var. 6 is the approach which only applies pose velocity.
\begin{table*}[!htb]
  \centering
  \caption{Ablations of the proposed method for short-term prediction. The six ablated approaches are numbered as 1, 2, ..., 6.}
    \begin{tabular}{l|c|c|c|cccc|cccc|cccc|cccc}
    \thickhline
 & VEL & POS & QT & \multicolumn{4}{c|}{Walking}  & \multicolumn{4}{c|}{Eating}  & \multicolumn{4}{c|}{Smoking} & \multicolumn{4}{c}{Discussion} \\
 \hline
   \multicolumn{4}{c|}{Milliseconds} &  80  & 160 & 320   & 400  & 80   & 160   & 320   & 400   & 80    & 160   & 320   & 400   & 80   & 160  & 320  & 400 \\
 \hline
& \checkmark & \checkmark & \checkmark & \textbf{0.20} & \textbf{0.35} & \textbf{0.54} & \textbf{0.59} & \textbf{0.18} & \textbf{0.32} & 0.54  & \textbf{0.66} & \textbf{0.22} & \textbf{0.44} & \textbf{0.81} & \textbf{0.91} & \textbf{0.24} & \textbf{0.60} & \textbf{0.83} & \textbf{0.93} \\
\hline
\textcircled{\small{1}} &  &  \checkmark & \checkmark & 0.23  & 0.38  & 0.56  & 0.62  & 0.21  & 0.34  & \textbf{0.53}  & 0.68  & 0.27  & 0.52  & 0.90  & 1.00  & 0.32  & 0.70  & 0.95  & 1.06 \\
\textcircled{\small{2}} & \checkmark  &  &  \checkmark & 0.21  & 0.35  & 0.54  & 0.59  & 0.21  & 0.35  & 0.55  & 0.68  & 0.25  & 0.50  & 0.88  & 0.97  & 0.27  & 0.64  & 0.91  & 0.98 \\
\textcircled{\small{3}} & \checkmark & \checkmark & & 0.23  & 0.39  & 0.59  & 0.66  & 0.21  & 0.36  & 0.60  & 0.73  & 0.28  & 0.53  & 0.90  & 1.03  & 0.30  & 0.70  & 1.03  & 1.15 \\
\textcircled{\small{4}} & & & \checkmark  & 0.24  & 0.40  & 0.62  & 0.69  & 0.23  & 0.40  & 0.65  & 0.79  & 0.29  & 0.56  & 0.98  & 1.11  & 0.31  & 0.68  & 0.98  & 1.08 \\
\textcircled{\small{5}} & & \checkmark & & 0.25  & 0.42  & 0.59  & 0.68  & 0.22  & 0.38  & 0.62  & 0.78  & 0.30  & 0.56  & 0.96  & 1.09  & 0.33  & 0.71  & 0.97  & 1.07 \\
\textcircled{\small{6}} & \checkmark &  & & 0.24  & 0.42  & 0.62  & 0.68  & 0.24  & 0.41  & 0.64  & 0.79  & 0.34  & 0.65  & 1.13  & 1.29  & 0.36  & 0.78  & 1.16  & 1.27 \\
    \thickhline
    \end{tabular} 
  \label{tab:ablation}
\end{table*}
\begin{table*}[!htb]
  \centering
  \caption{Ablations for the proposed method for long-term prediction on the Human3.6M dataset.}
    \begin{tabular}{l|c|c|c|cc|cc|cc|cc}
  \thickhline
  & VEL & POS & QT & \multicolumn{2}{c|}{Walking} & \multicolumn{2}{c|}{Eating} & \multicolumn{2}{c|}{Smoking} & \multicolumn{2}{c}{Discussion} \\
  \hline
    \multicolumn{4}{c|}{Milliseconds} & 560   & 1000  & 560   & 1000  & 560   & 1000  & 560   & 1000 \\
   \hline
    & \checkmark & \checkmark & \checkmark & \textbf{0.65} & \textbf{0.66} & \textbf{0.76} & \textbf{1.14}  & \textbf{0.97} & \textbf{1.42} & \textbf{1.29} & 1.77 \\
    \hline
    \textcircled{\small{1}} &  &  \checkmark & \checkmark &  0.70  & 0.71  & 0.83  & 1.19  & 1.09  & 1.56  & 1.44  & 1.98 \\
    \textcircled{\small{2}} & \checkmark  &  &  \checkmark & 0.65  & 0.68  & 0.80  & 1.12  & 1.03  & 1.51  & 1.32  & \textbf{1.72} \\
    \textcircled{\small{3}} & \checkmark & \checkmark & & 0.74  & 0.81  & 0.87  & 1.22  & 1.14  & 1.64  & 1.50  & 1.79 \\
    \textcircled{\small{4}} & & & \checkmark  & 0.75  & 0.85  & 0.93  & 1.30  & 1.24  & 1.80  & 1.47  & 1.98 \\
    \textcircled{\small{5}} & & \checkmark & &  0.77  & 0.84  & 0.90  & 1.29  & 1.21  & 1.67  & 1.43  & 1.81 \\
    \textcircled{\small{6}} & \checkmark &  & & 0.77  & 0.87  & 0.95  & 1.31  & 1.47  & 1.98  & 1.62  & 1.93 \\
  \thickhline
    \end{tabular} 
  \label{tab:ablation_long}
\end{table*}

\noindent \textbf{Pose Velocity}. Comparing our approach with Var. 1, we find that without velocities, the errors increase dramatically for both short-term predictions and long-term predictions. For example, velocity decreases the error by 0.05 for smoking and 0.08 for discussion at 80 milliseconds. It also decreases the error by 0.14 for smoking and 0.21 for discussion at 1000 milliseconds. While comparing Var. 3 and Var. 5, similar conclusions are reached for both periodic actions and aperiodic actions. The results are consistent with our hypothesis in Section~\ref{sec:pos_vel_rnn} that velocity preserves motion continuities and the input of velocity of human motion helps the network predict more accurate potential future poses.

\noindent \textbf{Position Embedding}. We compare our approach with Var. 2 to examine the effect of position embedding. We find that position embedding improves the results of prediction, and the improvement is even significant for aperiodic actions. For example, position embedding decreases the error by 0.03 at 80 milliseconds for eating, smoking and discussion. It also decreases the error by 0.06 for smoking at 560 milliseconds. We also find Var. 3 shows a substantial decreased error when compared with Var. 6. The results confirm our hypothesis in Section~\ref{sec:pos_emb} that positional information allows the network to learn more discriminative representations. 

\noindent \textbf{Quaternion Transformation}. To analyze the effect of quaternion transformation, we compare our approach with Var. 3 in Table~\ref{tab:ablation} and Table~\ref{tab:ablation_long} for both short-term prediction and long-term prediction. We find that quaternion transformation significantly contributes to human motion prediction. For example, the QT layer decreases the error by 0.06 for both smoking and discussion at 80 milliseconds. It also decreases the error by 0.22 for smoking at 1000 milliseconds. This significant improvement confirms the benefits of quaternion parameterization which is exempt from singularities and discontinuities (see Section~\ref{QuatTransf}).

\section{Conclusion}
This paper presents an end-to-end Position-Velocity Recurrent Encoder-Decoder (PVRED) for modeling and predicting human motion dynamics. PVRED incorporates pose velocity, position embedding and quaternion parameterization of human pose into a trainable network and learns to predict future poses based on a sequence of observed frames. Comprehensive experiments show that PVRED outperforms the state-of-the-art approaches for both short-term prediction and long-term prediction. Specifically, PVRED could generate human-like and meaningful poses in the future 4,000 milliseconds after the seed motion of 1,000 milliseconds. Further ablation studies validate the effects of each novel component of PVRED.

%
\IEEEpeerreviewmaketitle

\ifCLASSOPTIONcaptionsoff
  \newpage
\fi

\bibliographystyle{IEEEtran}
\bibliography{egbib}

\begin{thebibliography}{10}
\providecommand{\url}[1]{#1}
\csname url@samestyle\endcsname
\providecommand{\newblock}{\relax}
\providecommand{\bibinfo}[2]{#2}
\providecommand{\BIBentrySTDinterwordspacing}{\spaceskip=0pt\relax}
\providecommand{\BIBentryALTinterwordstretchfactor}{4}
\providecommand{\BIBentryALTinterwordspacing}{\spaceskip=\fontdimen2\font plus
\BIBentryALTinterwordstretchfactor\fontdimen3\font minus
  \fontdimen4\font\relax}
\providecommand{\BIBforeignlanguage}[2]{{%
\expandafter\ifx\csname l@#1\endcsname\relax
\typeout{** WARNING: IEEEtran.bst: No hyphenation pattern has been}%
\typeout{** loaded for the language `#1'. Using the pattern for}%
\typeout{** the default language instead.}%
\else
\language=\csname l@#1\endcsname
\fi
#2}}
\providecommand{\BIBdecl}{\relax}
\BIBdecl

\bibitem{ayusawa2017motion}
K.~Ayusawa and E.~Yoshida, ``Motion retargeting for humanoid robots based on
  simultaneous morphing parameter identification and motion optimization,''
  \emph{IEEE Transactions on Robotics}, vol.~33, no.~6, pp. 1343--1357, 2017.

\bibitem{villegas2018neural}
R.~Villegas, J.~Yang, D.~Ceylan, and H.~Lee, ``Neural kinematic networks for
  unsupervised motion retargetting,'' in \emph{Proceedings of the IEEE
  Conference on Computer Vision and Pattern Recognition}, 2018, pp. 8639--8648.

\bibitem{holden2016deep}
D.~Holden, J.~Saito, and T.~Komura, ``A deep learning framework for character
  motion synthesis and editing,'' \emph{ACM Transactions on Graphics}, vol.~35,
  no.~4, pp. 1--11, 2016.

\bibitem{martinez2017human}
J.~Martinez, M.~J. Black, and J.~Romero, ``On human motion prediction using
  recurrent neural networks,'' in \emph{IEEE Conference on Computer Vision and
  Pattern Recognition}.\hskip 1em plus 0.5em minus 0.4em\relax IEEE, 2017, pp.
  4674--4683.

\bibitem{brand2000style}
M.~Brand and A.~Hertzmann, ``Style machines,'' in \emph{Conference on Computer
  Graphics and Interactive Techniques}.\hskip 1em plus 0.5em minus 0.4em\relax
  ACM Press/Addison-Wesley Publishing Co., 2000, pp. 183--192.

\bibitem{pavlovic2001learning}
V.~Pavlovic, J.~M. Rehg, and J.~MacCormick, ``Learning switching linear models
  of human motion,'' in \emph{Advances In Neural Information Processing
  Systems}, 2001, pp. 981--987.

\bibitem{taylor2007modeling}
G.~W. Taylor, G.~E. Hinton, and S.~T. Roweis, ``Modeling human motion using
  binary latent variables,'' in \emph{Advances in Neural Information Processing
  Systems}, 2007, pp. 1345--1352.

\bibitem{fragkiadaki2015recurrent}
K.~Fragkiadaki, S.~Levine, P.~Felsen, and J.~Malik, ``Recurrent network models
  for human dynamics,'' in \emph{IEEE International Conference on Computer
  Vision}.\hskip 1em plus 0.5em minus 0.4em\relax IEEE, 2015, pp. 4346--4354.

\bibitem{jain2016structural}
A.~Jain, A.~R. Zamir, S.~Savarese, and A.~Saxena, ``Structural-{RNN}: {D}eep
  learning on spatio-temporal graphs,'' in \emph{IEEE Conference on Computer
  Vision and Pattern Recognition}.\hskip 1em plus 0.5em minus 0.4em\relax IEEE,
  2016, pp. 5308--5317.

\bibitem{ghosh2017learning}
P.~Ghosh, J.~Song, E.~Aksan, and O.~Hilliges, ``Learning human motion models
  for long-term predictions,'' in \emph{International Conference on 3D
  Vision}.\hskip 1em plus 0.5em minus 0.4em\relax IEEE, 2017, pp. 458--466.

\bibitem{vaswani2017attention}
A.~Vaswani, N.~Shazeer, N.~Parmar, J.~Uszkoreit, L.~Jones, A.~N. Gomez,
  {\L}.~Kaiser, and I.~Polosukhin, ``Attention is all you need,'' in
  \emph{Advances in Neural Information Processing Systems}, 2017, pp.
  5998--6008.

\bibitem{gehring2017convolutional}
J.~Gehring, M.~Auli, D.~Grangier, D.~Yarats, and Y.~N. Dauphin, ``Convolutional
  sequence to sequence learning,'' in \emph{International Conference on Machine
  Learning}.\hskip 1em plus 0.5em minus 0.4em\relax JMLR, 2017, pp. 1243--1252.

\bibitem{grassia1998practical}
F.~S. Grassia, ``Practical parameterization of rotations using the exponential
  map,'' \emph{Journal of Graphics Tools}, vol.~3, no.~3, pp. 29--48, 1998.

\bibitem{pavllo:quaternet:2018}
D.~Pavllo, D.~Grangier, and M.~Auli, ``{QuaterNet}: {A} quaternion-based
  recurrent model for human motion,'' in \emph{British Machine Vision
  Conference}.\hskip 1em plus 0.5em minus 0.4em\relax BMVA, 2018, pp. 675--688.

\bibitem{tang2018long}
Y.~Tang, L.~Ma, W.~Liu, and W.~Zheng, ``Long-term human motion prediction by
  modeling motion context and enhancing motion dynamic,'' in
  \emph{International Joint Conference on Artificial Intelligence}, 2018, pp.
  935--941.

\bibitem{chiu2018action}
H.-k. Chiu, E.~Adeli, B.~Wang, D.-A. Huang, and J.~C. Niebles,
  ``Action-agnostic human pose forecasting,'' in \emph{Winter Conference on
  Applications of Computer Vision}.\hskip 1em plus 0.5em minus 0.4em\relax
  IEEE, 2018, pp. 1423--1432.

\bibitem{li2018convolutional}
C.~Li, Z.~Zhang, W.~S. Lee, and G.~H. Lee, ``Convolutional sequence to sequence
  model for human dynamics,'' in \emph{IEEE Conference on Computer Vision and
  Pattern Recognition}.\hskip 1em plus 0.5em minus 0.4em\relax IEEE, 2018, pp.
  5226--5234.

\bibitem{cho2014learning}
K.~Cho, B.~van Merrienboer, C.~Gulcehre, D.~Bahdanau, F.~Bougares, H.~Schwenk,
  and Y.~Bengio, ``Learning phrase representations using rnn encoder--decoder
  for statistical machine translation,'' in \emph{Empirical Methods in Natural
  Language Processing}, 2014, pp. 1724--1734.

\bibitem{sutskever2014sequence}
I.~Sutskever, O.~Vinyals, and Q.~V. Le, ``Sequence to sequence learning with
  neural networks,'' in \emph{Advances in Neural Information Processing
  Systems}, 2014, pp. 3104--3112.

\bibitem{gui2018adversarial}
L.-Y. Gui, Y.-X. Wang, X.~Liang, and J.~M. Moura, ``Adversarial geometry-aware
  human motion prediction,'' in \emph{European Conference on Computer
  Vision}.\hskip 1em plus 0.5em minus 0.4em\relax Springer, 2018, pp. 823--842.

\bibitem{gopalakrishnan2018a}
A.~Gopalakrishnan, A.~Mali, D.~Kifer, C.~L. Giles, and A.~G. Ororbia, ``A
  neural temporal model for human motion prediction.'' in \emph{IEEE Conference
  on Computer Vision and Pattern Recognition}.\hskip 1em plus 0.5em minus
  0.4em\relax IEEE, 2019, pp. 12\,116--12\,125.

\bibitem{butepage2017deep}
J.~B{\"u}tepage, M.~J. Black, D.~Kragic, and H.~Kjellstr{\"o}m, ``Deep
  representation learning for human motion prediction and classification,'' in
  \emph{Conference on Computer Vision and Pattern Recognition}.\hskip 1em plus
  0.5em minus 0.4em\relax IEEE, 2017, pp. 6158--6166.

\bibitem{ruiz2018human}
A.~H. Ruiz, J.~Gall, and F.~Morenonoguer, ``Human motion prediction via
  spatio-temporal inpainting.'' in \emph{IEEE International Conference on
  Computer Vision}.\hskip 1em plus 0.5em minus 0.4em\relax IEEE, 2019, pp.
  7134--7143.

\bibitem{wang2019imitation}
B.~Wang, E.~Adeli, H.~Chiu, D.~Huang, and J.~C. Niebles, ``Imitation learning
  for human pose prediction,'' in \emph{IEEE International Conference on
  Computer Vision}.\hskip 1em plus 0.5em minus 0.4em\relax IEEE, 2019, pp.
  7124--7133.

\bibitem{Aksan_2019_ICCV}
E.~Aksan, M.~Kaufmann, and O.~Hilliges, ``Structured prediction helps 3d human
  motion modelling,'' in \emph{IEEE International Conference on Computer
  Vision}.\hskip 1em plus 0.5em minus 0.4em\relax IEEE, 2019, pp. 7144--7153.

\bibitem{wei2019motion}
M.~Wei, L.~Miaomiao, S.~Mathieu, and L.~Hongdong, ``Learning trajectory
  dependencies for human motion prediction,'' in \emph{IEEE International
  Conference on Computer Vision}, 2019, pp. 9489--9497.

\bibitem{aksan2020spatio}
E.~Aksan, P.~Cao, M.~Kaufmann, and O.~Hilliges, ``A spatio-temporal transformer
  for 3d human motion prediction,'' \emph{2020, arXiv:2004.08692. [Online].
  Available: http://arxiv.org/abs/2004.08692}.

\bibitem{li2020dynamic}
M.~Li, S.~Chen, Y.~Zhao, Y.~Zhang, Y.~Wang, and Q.~Tian, ``Dynamic multiscale
  graph neural networks for 3d skeleton based human motion prediction,'' in
  \emph{IEEE Conference on Computer Vision and Pattern Recognition}, 2020, pp.
  214--223.

\bibitem{lehrmann2014efficient}
A.~M. Lehrmann, P.~V. Gehler, and S.~Nowozin, ``Efficient nonlinear markov
  models for human motion,'' in \emph{IEEE Conference on Computer Vision and
  Pattern Recognition}.\hskip 1em plus 0.5em minus 0.4em\relax IEEE, 2014, pp.
  1314--1321.

\bibitem{wang2017modeling}
H.~Wang and L.~Wang, ``Modeling temporal dynamics and spatial configurations of
  actions using two-stream recurrent neural networks,'' in \emph{IEEE
  Conference on Computer Vision and Pattern Recognition}.\hskip 1em plus 0.5em
  minus 0.4em\relax IEEE, 2017, pp. 499--508.

\bibitem{wang2018beyond}
------, ``Beyond joints: {L}earning representations from primitive geometries
  for skeleton-based action recognition and detection,'' \emph{IEEE
  Transactions on Image Processing}, vol.~27, no.~9, pp. 4382--4394, 2018.

\bibitem{sidenbladh2002implicit}
H.~Sidenbladh, M.~J. Black, and L.~Sigal, ``Implicit probabilistic models of
  human motion for synthesis and tracking,'' in \emph{European Conference on
  Computer Vision}.\hskip 1em plus 0.5em minus 0.4em\relax Springer, 2002, pp.
  784--800.

\bibitem{lehrmann2013non}
A.~M. Lehrmann, P.~V. Gehler, and S.~Nowozin, ``A non-parametric bayesian
  network prior of human pose,'' in \emph{IEEE International Conference on
  Computer Vision}.\hskip 1em plus 0.5em minus 0.4em\relax IEEE, 2013, pp.
  1281--1288.

\bibitem{wang2006gaussian}
J.~Wang, A.~Hertzmann, and D.~J. Fleet, ``Gaussian process dynamical models,''
  in \emph{Advances in Neural Information Processing Systems}, 2006, pp.
  1441--1448.

\bibitem{taylor2009factored}
G.~W. Taylor and G.~E. Hinton, ``Factored conditional restricted boltzmann
  machines for modeling motion style,'' in \emph{International Conference on
  Machine Learning}.\hskip 1em plus 0.5em minus 0.4em\relax ACM, 2009, pp.
  1025--1032.

\bibitem{holden2015learning}
D.~Holden, J.~Saito, T.~Komura, and T.~Joyce, ``Learning motion manifolds with
  convolutional autoencoders,'' in \emph{SIGGRAPH Asia Technical Briefs}.\hskip
  1em plus 0.5em minus 0.4em\relax ACM, 2015, pp. 1--4.

\bibitem{hochreiter1997long}
S.~Hochreiter and J.~Schmidhuber, ``Long short-term memory,'' \emph{Neural
  Computation}, vol.~9, no.~8, pp. 1735--1780, 1997.

\bibitem{ionescu2014human3}
C.~Ionescu, D.~Papava, V.~Olaru, and C.~Sminchisescu, ``Human3. 6m: {L}arge
  scale datasets and predictive methods for 3d human sensing in natural
  environments,'' \emph{IEEE Transactions on Pattern Analysis and Machine
  Intelligence}, vol.~36, no.~7, pp. 1325--1339, 2014.

\end{thebibliography}
\begin{IEEEbiography}[{\includegraphics[width=1in,height=1.25in]{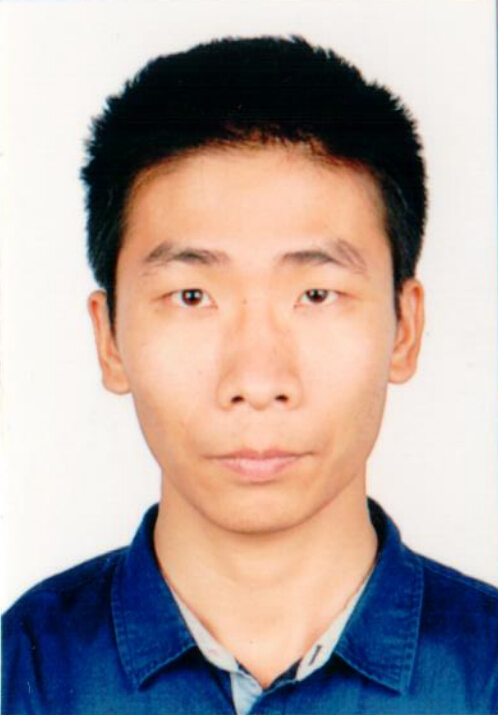}}]{Hongsong Wang}
	received B.E. degree in Automation from Huazhong University of Science and Technology in 2013 and Ph.D. degree in Pattern Recognition and Intelligent Systems from Institute of Automation, University of Chinese Academy of Sciences in 2018. He was a postdoctoral fellow at National University of Singapore in 2019. After that, he served as a research associate at Inception Institute of Artificial Intelligence, Abu Dhabi, UAE in 2020. Currently, he is a researcher at Tencent. His research interests include deep learning based applications such as pedestrian detection, video representation, action recognition, multi-object tracking and video instance segmentation.
\end{IEEEbiography}

\begin{IEEEbiography}[{\includegraphics[width=1in,height=1.25in,clip,keepaspectratio]{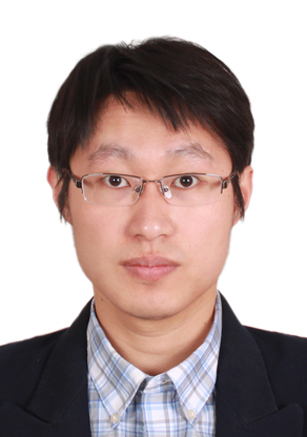}}]{Jian Dong}
	received his Ph.D. degree from National University of Singapore, and B.E. degree from the University of Science and Technology of China. He is currently a senior director with Qihoo 360. Prior to that, he was a research scientist with Amazon. His research interests include machine learning and computer vision. He received winner prizes in both PASCAL VOC and ILSVRC competitions.
\end{IEEEbiography}

\begin{IEEEbiography}[{\includegraphics[width=1in,height=1.25in,clip,keepaspectratio]{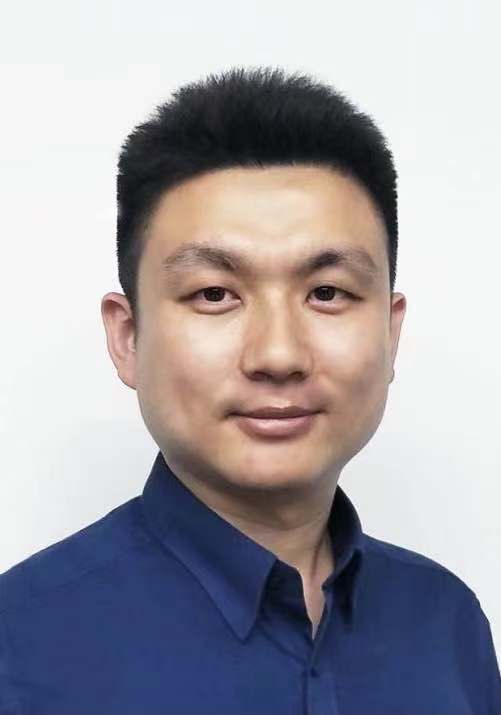}}]{Bin Cheng}
	is currently a research director in the Beijing Academy of Artificial Intelligence (BAAI). He received his Ph.D. degree from National University of Singapore, and B.E. degree from the University of Science and Technology of China. His research areas include computer vision, machine learning and the related applications. And he also shipped a dozen of technologies to many industry products, including smart devices, search, live stream, short video and financial risk-control.
\end{IEEEbiography}

\begin{IEEEbiography}[{\includegraphics[width=1in,height=1.25in,clip,keepaspectratio]{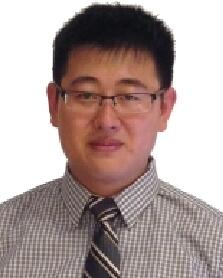}}]{Jiashi Feng}
	is currently an Assistant Professor in the Department of Electrical and Computer Engineering at National University of Singapore. He got his B.E. degree from University of Science and Technology, China in 2007 and Ph.D. degree from National University of Singapore in 2014. He was a postdoc researcher at University of California, Berkeley from 2014 to 2015. He received TR35 Asia in 2018, NUS Early Career Research Award and winner prize of object localization task of ILSVRC 2017. His current research interest focuses on machine learning and computer vision techniques for large-scale data analysis. Specifically, he has done work in object recognition, deep learning, machine learning, high-dimensional statistics and big data analysis.
\end{IEEEbiography}

\end{document}